\documentclass{article}

\usepackage{wrapfig}
\usepackage{microtype}
\usepackage{graphicx}
\usepackage{subfigure}
\usepackage{booktabs} 
\usepackage{amsmath,amssymb}
\usepackage{natbib}
\usepackage{caption}
\usepackage{tabularx} 
\usepackage{threeparttable}

\usepackage{diagbox}

\newtheorem{theorem}{Theorem}[section]
\newtheorem{corollary}{Corollary}[theorem]
\newtheorem{lemma}[theorem]{Lemma}

\usepackage{enumitem}
\setlist[itemize]{leftmargin=0.5cm}
\setlist[enumerate]{leftmargin=0.5cm}

\input{def.set}

\newcommand{\myendofproof}[0]{\hfill $\blacksquare$ \newline}

\usepackage{hyperref}

\newcommand{\modelname}[1]{$\text{TWIRLS}_\text{#1}$}

\usepackage[accepted]{arxivver}

\icmltitlerunning{Graph Neural Networks Inspired by Classical Iterative Algorithms}

\begin{document}

\twocolumn[
\icmltitle{Graph Neural Networks Inspired by Classical Iterative Algorithms}



\icmlsetsymbol{equal}{*}
\icmlsetsymbol{wkdaamz}{\dag}

\begin{icmlauthorlist}

\icmlauthor{Yongyi Yang}{fdu,wkdaamz}
\icmlauthor{Tang Liu}{fdu,wkdaamz} 
\icmlauthor{Yangkun Wang}{sjtu,wkdaamz}
\icmlauthor{Jinjing Zhou}{amz}
\icmlauthor{Quan Gan}{amz}
\icmlauthor{Zhewei Wei}{rmu}
\icmlauthor{Zheng Zhang}{amz}
\icmlauthor{Zengfeng Huang}{fdu}
\icmlauthor{David Wipf}{amz} 

\end{icmlauthorlist}

\icmlaffiliation{fdu}{Fudan University}
\icmlaffiliation{sjtu}{Shanghai Jiao Tong University}
\icmlaffiliation{rmu}{Renmin University of China}
\icmlaffiliation{amz}{Amazon}

\icmlcorrespondingauthor{Zengfeng Huang}{huangzf@fudan.edu.cn}
\icmlcorrespondingauthor{David Wipf}{davidwipf@gmail.com}

\icmlkeywords{Machine Learning}

\vskip 0.3in
]


\printAffiliationsAndNotice{\textsuperscript{\dag}Work completed during an internship at the AWS Shanghai AI Lab.}

\begin{abstract}
Despite the recent success of graph neural networks (GNN), common architectures often exhibit significant limitations, including sensitivity to oversmoothing, long-range dependencies, and spurious edges, e.g., as can occur as a result of graph heterophily or adversarial attacks.  To at least partially address these issues within a simple transparent framework, we consider a new family of GNN layers designed to mimic and integrate the update rules of two classical iterative algorithms, namely, proximal gradient descent and iterative reweighted least squares (IRLS).  The former defines an extensible base GNN architecture that is immune to oversmoothing while nonetheless capturing long-range dependencies by allowing arbitrary propagation steps.  In contrast, the latter produces a novel attention mechanism that is explicitly anchored to an underlying end-to-end energy function, contributing stability with respect to edge uncertainty.  When combined we obtain an extremely simple yet robust model that we evaluate across disparate scenarios including standardized benchmarks, adversarially-perturbated graphs, graphs with heterophily, and graphs involving long-range dependencies.  In doing so, we compare against SOTA GNN approaches that have been explicitly designed for the respective task, achieving competitive or superior node classification accuracy.  Our code is available at  \href{https://github.com/FFTYYY/TWIRLS/}{\underline{this link}}.
\end{abstract}

\vspace*{-0.5cm}
\section{Introduction} \label{sec:intro}
\vspace*{-0.1cm}
Given data comprised of entities with explicit relationships between them, graph neural networks (GNNs) represent a powerful means of exploiting these relationships within a predictive model \cite{DBLP:journals/corr/abs-1812-08434}.  Among the most successful candidates in this vein, message-passing GNNs are based on stacking a sequence of propagation layers, whereby neighboring nodes share feature information via an aggregation function \cite{kipf2017semi,HamiltonYL17,DBLP:journals/jcamd/KearnesMBPR16}.  This sharing is also at times modulated by various forms of attention to effectively compute propagation weights that reflect the similarity between nodes connected by an edge \cite{VelickovicCCRLB18}.

While such approaches display impressive performance on graph benchmarks involving tasks such as semi-supervised node classification conditioned on input features, non-trivial shortcomings remain.  For example, as we deviate from more standardized regimes exhibiting graph homophily, meaning nearby nodes share similar labels and features, into the less familiar, contrasting domain of heterophily, many message-passing GNNs (including those with attention) experience a steep degradation in accuracy \cite{DBLP:conf/nips/ZhuYZHAK20,DBLP:journals/corr/abs-2009-13566}.  Similarly, when edge uncertainty is introduced via adversarial attacks or related, performance declines as well \cite{DBLP:conf/ijcai/ZugnerAG19,DBLP:conf/iclr/ZugnerG19}.

Pushing further, for problems requiring that nodes share information across non-local regions of the graph, message-passing GNNs require a large number of propagation layers.  However, this can lead to a well-known oversmoothing phenomena in which node features converge to similar, non-discriminative embeddings \cite{DBLP:conf/iclr/OonoS20,DBLP:conf/aaai/LiHW18}.  To this end, a wide variety of countermeasures have been deployed to address this concern, most notably various types of skip connections designed to preserve local information even while propagating to distant neighbors \cite{DBLP:conf/icml/ChenWHDL20,DBLP:conf/iccv/Li0TG19,DBLP:conf/icml/XuLTSKJ18}. Even so, most deeper architectures generally display modest improvement, and may require additional tuning or regularization strategies to be effective.

And finally, beyond the above issues there is at times a lingering opaqueness in terms of why and how different GNN architectural choices relate to performance on specific tasks.  This is due, seemingly in part, to the importance of nuanced heuristics in obtaining SOTA performance on graph leaderboards, as opposed to more transparent designs that are likely to consistently transfer to new domains.


\textbf{Our Contributions} ~ As a partial step towards mitigating these concerns, we propose a transparent GNN architecture with an inductive bias inspired by classical iterative algorithms.  In doing so, our design benefits from the following:
\begin{itemize}
    \item All architectural components are in one-to-one correspondence with the unfolded iterations of robust descent algorithms applied to minimizing a principled graph-regularized energy function.  Specifically, we adopt propagation layers and nonlinear activations  patterned after proximal gradient updates (Section \ref{sec:base_Model}), while we integrate these steps with graph attention using iterative reweighted least squares (IRLS) (Section \ref{sec:IRLS}).  To the best of our knowledge, this is the first end-to-end GNN model, including attention, designed based on such an unfolding perspective (Section \ref{sec:related_work}).
    \item By anchoring to a principled energy function, we can include an arbitrary number of propagation layers to capture non-local dependencies, but with no risk of oversmoothing.  And by unifying each layer via IRLS reweighting functions, we can include a novel family of attention mechanisms that simultaneously contribute robustness to graph uncertainty without introducing additional parameters or design heuristics.
    \item In head-to-head comparisons with SOTA methods explicitly tailored for disparate testing conditions including standard benchmarks, graphs with heterophily, and adversarial perturbations, our proposed architecture achieves competitive or superior node classification accuracy (Section \ref{sec:experiments}).
\end{itemize}

\section{A Simple GNN Backbone Derived from Proximal Gradient Descent}\label{sec:base_Model}

Our starting point will be a class of affine GNN layers that are immune to oversmoothing and yet are sufficiently flexible to handle nonlinear activations, sampling, and layer-dependent weights within a unified framework.  As described next, this is achievable by unfolding the iterations of proximal gradient descent methods applied to a suitably regularized energy function.  

\subsection{Extensible GNN Layers via Unfolding} \label{eq:basic_unfolding}

Given a graph $\calG = \{\calV,\calE\}$, with $n = |\calV|$ nodes and $m=|\calE|$ edges, inspired by \cite{zhou2004learning} we define the energy function
\begin{equation} \label{eq:basic_objective}
\ell_Y(Y) \triangleq  \left\|Y - f\left(X ; W  \right) \right\|_{\calF}^2 + \lambda \mbox{tr}\left[Y^\top L Y  \right],
\end{equation}
where $\lambda$ is a trade-off parameter and $Y \in \mathbb{R}^{n\times d}$ represents a candidate embedding of $d$-dimensional features across $n$ nodes.  Similarly, $f\left(X;W\right)$ denotes a function (parameterized by $W$) of initial $d_0$-dimensional node features $X \in \mathbb{R}^{n\times d_0}$.  For example, we can have $f\left(X;W\right) = X W$ or $f\left(X;W\right) = \mbox{MLP}\left[X; W \right]$, i.e., a multi-layer perceptron with weights $W$.  And finally, $L \in \mathbb{R}^{n\times n}$ is the Laplacian of $\calG$, meaning $L = D-A = B^\top B$, where $D$ and $A$ are degree and adjacency matrices respectively, while $B \in \mathbb{R}^{m\times n}$ is an incidence matrix.

It is not difficult to show that
\begin{equation} \label{eq:optimal_L2_solution}
Y^*\left( W \right) \triangleq \arg \min_{Y} \ell_Y(Y) = \left(I +  \lambda L \right)^{-1} f\left(X ; W  \right),
\end{equation}
where obviously the optimal $Y^*$ will be a function of $W$. This solution represents a loose approximation to $f\left(X ; W  \right)$ that has been smoothed across the graph structure via the trace term in (\ref{eq:basic_objective}), balancing local and global information  \cite{zhou2004learning}.  Hence it is reasonable to treat $Y^*\left( W \right)$ as graph-aware features for application to a downstream prediction problem of interest, e.g., node classification.

Given that $Y^*\left( W \right)$ is differentiable w.r.t.~$W$, we need only plug into a node classification or regression loss of the form
\begin{equation} \label{eq:meta_loss}
\ell_W\left(W, \theta \right) \triangleq \sum_{i=1}^{n'} \calD\bigg( g\left[ \by_{i}^*\left( W \right); \theta \right], t_i  \bigg),
\end{equation}
where $g : \mathbb{R}^d \rightarrow \mathbb{R}$ is a node-wise function with parameters $\theta$, e.g., $g$ could be a linear transformation, an MLP, or even an identity mapping.  Additionally, $\by_{i}^*\left(W\right)$ is the $i$-th row of $Y^*\left( W \right)$, $t_i$ is a ground-truth target label of node $i$, and $\calD$ is some discriminator function, e.g., cross-entropy for classification, squared error for regression.\footnote{Note that without loss of generality, (\ref{eq:meta_loss}) implicitly assumes the first $n'<n$ nodes have been labeled for training.} We may then optimize $\ell_W\left(W, \theta \right)$ over both $W$ and $\theta$.  In this capacity, each $\by_{i}^*\left(W\right)$ serves as a trainable node-level feature that has been explicitly regularized via the graph structure.

While the above strategy may be feasible for small graphs, if $n$ is large then computing the inverse of $I + \lambda L$, a requirement for producing $Y^*(W)$ analytically, can be prohibitively expensive.  As a more feasible alternative, we may instead approximate $Y^*(W)$ with an estimate formed by taking gradient steps along (\ref{eq:basic_objective}) with respect to $Y$, starting from some initial point $Y^{(0)}$, e.g., $Y^{(0)} = f\left(X ; W  \right)$ \cite{zhou2004learning}.  As long as these gradient steps are themselves differentiable w.r.t.~$W$, we can still plug the resulting approximate estimator $\hat{Y}(W)$ into the meta-objective (\ref{eq:meta_loss}) and proceed as before.  In this sense we are essentially \textit{unfolding} a sequence of gradient steps to form a differentiable chain that can be viewed as trainable layers of a deep architecture \cite{gregor2010learning,hershey2014deep,Sprechmann15}.  Analogous strategies have been used in the past to facilitate meta-learning using features originating from a lower-level loss whose optimum itself is not directly differentiable \cite{wang2016learning}.

To proceed, we note that
\begin{equation} 
\frac{\partial \ell_Y(Y) }{\partial Y} = 2\lambda L Y + 2Y - 2 f\left(X ; W  \right),
\end{equation}
and therefore iteration $k+1$ of gradient descent can be computed as
\begin{equation} \label{eq:basic_grad_step}
Y^{(k+1)} = Y^{(k)} - \alpha\left[ Q Y^{(k)} - f\left(X ; W  \right) \right], ~~Q \triangleq \lambda L  + I,
\end{equation}
where $\tfrac{\alpha}{2}$ is the step size.  But if $Q$ has a large condition number, the gradient descent convergence rate can be slow \cite{nocedal2006numerical}.  Fortunately though, preconditioning techniques exist to reduce the convergence time.



For example, as motivated by the Richardson iterative method for solving linear systems \cite{axelsson1996iterative}, Jacobi preconditioning involves simply rescaling each gradient step using $\left( \mbox{diag}[Q] \right)^{-1} = \left(\lambda D + I\right)^{-1}$.  After rearranging terms and defining the diagonal matrix $\tilde{D} \triangleq \lambda D+I$, this leads to the modified update
\begin{equation} \label{eq:basic_grad_step2}
Y^{(k+1)} = (1-\alpha) Y^{(k)} + \alpha \tilde{D}^{-1} \left[ \lambda A Y^{(k)} + f\left(X ; W  \right) \right].
\end{equation}
We will henceforth treat this expression as the $k$-th differentiable propagation layer of a GNN model; later we will discuss enhancements of this baseline designed to provide greater flexibility and robustness.  Note also that, scale factors $(1-\alpha)$ and $\alpha \tilde{D}^{-1}$ notwithstanding, (\ref{eq:basic_grad_step2}) can be interpreted as having skip connections from both the previous layer $Y^{(k)}$ and the input features $f\left(X ; W  \right)$.  While prior work has also advocated the use of skip connections in various forms to help mitigate oversmoothing effects \cite{DBLP:conf/icml/ChenWHDL20,DBLP:conf/iccv/Li0TG19,DBLP:conf/icml/XuLTSKJ18}, within the unfolding framework such connections naturally emerge from an underlying energy function, i.e., they are not introduced as a post hoc patch.

\subsection{Relationship with Graph Convolution Layers}

Interestingly, under suitable assumptions detailed in the supplementary, if we choose $Y^{(0)} = f\left(X ; W  \right) = \tilde{D}^{1/2} X W$, then the first gradient step $Y^{(1)}$ is equivalent, up to a simple reparameterization by $\tilde{D}^{1/2}$, to the \textit{graph convolutional network} (GCN) layer $\tilde{D}^{-1/2} \tilde{A}\tilde{D}^{-1/2} X W$, where $\tilde A = A + I$.  See \cite{kipf2017semi}  for further details regarding the GCN architecture.  We now examine similarities and differences across arbitrary iterations with respect to oversmoothing, nonlinear activations, layer-dependent weights, and sampling.

\textbf{Oversmoothing}~  Unlike $k=1$, for iterations $k>1$, (\ref{eq:basic_grad_step2}) diverges significantly from GCN layers.  This is reflected in critical differences in the asymptotic propagation of (\ref{eq:basic_grad_step2}) versus the analogous GCN-based rule 
\begin{equation} \label{eq:SGC}
Y^{(k+1)} = \tilde{D}^{-1/2} \tilde{A}\tilde{D}^{-1/2} Y^{(k)} W,
\end{equation}
where (\ref{eq:SGC}) has been explored in \cite{wu2019simplifying} under the moniker \textit{simple graph convolutions} (SGC).  Specifically, while (\ref{eq:basic_grad_step2}) converges to the solution given by (\ref{eq:optimal_L2_solution}), which represents a principled graph diffusion akin to personalized page rank \cite{klicpera2019predict}, SGC with $k \rightarrow \infty$ converges to an \textit{oversmoothed} solution whereby $\by^{(\infty)}_{i} = \bc$ for all $i \in \calV$, i.e., all node embeddings converge to a constant $\bc$ \cite{wu2019simplifying}.  As such, by treating (\ref{eq:basic_grad_step2}) as the propagation rule for GNN layers, oversmoothing is not a concern.  This is worth noting given the signification consideration given to this topic in the literature \cite{DBLP:conf/iclr/OonoS20,DBLP:conf/aaai/LiHW18,DBLP:conf/iclr/RongHXH20}.

\textbf{Nonlinear Activations}~ The original GCN model from \cite{kipf2017semi} followed each convolutional layer with a node-wise nonlinear ReLU activation function.  Nonlinear activations can also naturally be introduced within the unfolding framework described in Section \ref{eq:basic_unfolding}.  In fact, there is a direct correspondence between the inclusion of an additional nonlinear regularization term (or constraint) in (\ref{eq:basic_objective}), and the introduction of a nonlinear activation within the update (\ref{eq:basic_grad_step2}).  Specifically, let $\eta : \mathbb{R}^{d} \rightarrow \mathbb{R}^{d}$ denote any function of individual node features.  The revised optimization problem
\begin{equation}
\min_Y ~ \ell_Y(Y) + \sum_{i} \eta\left(\by_{i}\right)
\end{equation}
can be efficiently solved using proximal gradient methods \cite{combettes2011proximal}, a generalized form of projected gradient descent, whenever the proximal operator
\begin{equation}
    \mbox{prox}_{ \eta}\left( \bu \right) \triangleq \arg\min_{\by} \tfrac{1}{2}\left\|\bu -\by \right\|_2^2 + \eta\left( \by \right) 
\end{equation}
is easily computable in closed form when applied to any generic input argument $\bu$. As an illustrative special case, if $\eta(\by) = \sum_j \calI_{\infty}[y_j < 0]$, meaning an indicator function assigning infinite penalty to any $y_j < 0$ (this is effectively equivalent to a constraint on $\by$ to the positive orthant), the corresponding proximal operator satisfies $\mbox{prox}_{\eta}\left( \bu \right) = \mbox{ReLU}(\bu) = \max\left({\bf 0},\bu \right).$
The proximal descent version of (\ref{eq:basic_grad_step2}) then becomes
\begin{eqnarray} \label{eq:basic_grad_step3}
&& \hspace*{-0.7cm} Y^{(k+1)} = \\
&& \hspace*{-0.4cm} \mbox{ReLU}\left(  (1-\alpha) Y^{(k)} + \alpha \tilde{D}^{-1} \left[ \lambda A Y^{(k)} + f\left(X ; W  \right) \right] \right), \nonumber 
\end{eqnarray}
with guaranteed descent for suitable step sizes \cite{combettes2011proximal}.  In wider contexts, proximal gradient methods have been applied to designing deep architectures for structured regression tasks \cite{Sprechmann15}. 



\textbf{Layer-Dependent Weights and Sampling}~ Within the unfolding framework, layer-dependent weights can be effectively introduced by simply altering the specification of the norms used to define $\ell_Y(Y)$ such that a parameterized/learnable metric is included.  Additionally, sampling methods designed to sparsify the graph, which can significantly reduce time and memory costs \cite{chen2018fastgcn}, can also naturally be accommodated using an unbiased stochastic estimate of $\ell_Y(Y)$.  In both cases, please see the supplementary for more details. 


\section{Robust Regularization Using IRLS-Based Graph Attention} \label{sec:IRLS}
Now suppose we do \textit{not} believe that 
\vspace*{-0.0cm}
\begin{equation} \label{L2_norm_penalty}
\mbox{tr}\left[Y^\top L Y  \right] = \| B Y \|^2_{\calF} = \sum_{\{i,j\} \in \calE} \left\| \by_{i} - \by_{j} \right\|_2^2
\end{equation}
is the most appropriate regularization factor for incorporating graph structure within (\ref{eq:basic_objective}) (here we are assuming that all edge weights are unity, although this assumption can be relaxed). To inject additional flexibility, we may consider upgrading (\ref{L2_norm_penalty}),  leading to the revised energy  $\ell_Y(Y; \rho) \triangleq$
\begin{equation} \label{eq:robust_objective}
\left\|Y - f\left(X ; W  \right) \right\|_{\calF}^2 + \lambda \sum_{\{i,j\} \in \calE} \rho\left( \left\| \by_{i} - \by_{j} \right\|_2^2 \right).
\end{equation}
Here $\rho : \mathbb{R}^+ \rightarrow \mathbb{R}$ is some nonlinear function designed to contribute robustness, particularly w.r.t~edge uncertainty. 

In this section, we will first motivate how edge uncertainty, which undercuts the effectiveness of (\ref{L2_norm_penalty}), can be modeled using Gaussian scale mixtures (GSM) \cite{andrews1974scale}.  We later apply this perspective to informing suitable choices for $\rho$ in (\ref{eq:robust_objective}).  Finally, we map the updated energy to an idealized graph attention mechanism that can be integrated within the unfolding framework from Section \ref{sec:base_Model} via IRLS \cite{Chartrand08,daubechies2010iteratively}.

\subsection{Edge Uncertainty and Gaussian Scale Mixtures}
The quadratic penalty (\ref{L2_norm_penalty}) is in some sense the optimal regularization factor for Gaussian errors aligned with the graph structure \cite{jia2020residual}, and indeed, when we apply a $\exp\left[-\left(\cdot\right)\right]$ transformation and suitable normalization we obtain a structured Gaussian prior distribution $p(Y)$ with unit variance along each edge.  In fact, from this perspective minimizing (\ref{eq:basic_objective}) is equivalent to \textit{maximum a posteriori} (MAP) estimation via $p(Y|X) \propto p(X|Y)p(Y)$, with likelihood $p(X|Y) \propto \exp\left[-\tfrac{1}{2\lambda}\left\|Y - f\left(X ; W  \right) \right\|_{\calF}^2\right]$.

But of course it is well known that $\ell_2$-norms and Gaussian priors are very sensitive to outliers, since errors accumulate quadratically \cite{west1984outlier}.  In the present context, this would imply that spurious edges (e.g.,  that were errantly included in the graph) or which link nodes with neither label nor features in common, could dominate the resulting loss leading to poor performance on downstream tasks.

Consequently, given the potential weakness of (\ref{L2_norm_penalty}), and by extension (\ref{eq:basic_objective}), it may be preferable to introduce some uncertainty into the allowable variance/penalty along each edge.  More specifically, we can replace the fixed, unit variances implicitly assumed above with the more flexible Gaussian scale mixture prior defined as
\begin{equation} \label{eq:sparse_graph_prior}
p\left(Y \right) = Z^{-1} \prod_{\{i,j\}\in \calE} \int  \calN\left(\bu_{ij} |  0,\gamma^{-1}_{ij} I \right) d \mu\left(\gamma_{ij}\right),
\end{equation}
where $\mu$ is some positive measure over latent precision parameters $\gamma_{ij}$, $\bu_{ij} \triangleq  \by_{i} - \by_{j} ~ \forall \{i,j\}\in \calE$, and $Z$ is a standard partition function that ensures (\ref{eq:sparse_graph_prior}) sums to one.  When the measure $\mu$ assigns all of its mass to $\gamma_{ij} = 1$ for all edges, then $-\log p\left(Y\right)$ so-defined collapses to (\ref{L2_norm_penalty}); however, in broader scenarios this model allows us to consider a distribution of precision (or variance) across edges.  For example, if $\mu$ allocates non-trivial mass to small precision values, it reflects our belief that some edges may be spurious, and the penalty on large values of $\| \by_{i} - \by_{j} \|_2$ is reduced (i.e., it is rescaled by $\gamma_{ij}$).  As a special case, neighbor sampling \cite{chen2018fastgcn} can be motivated by choosing $\mu$ such that all probability mass is partitioned between $\gamma_{ij} = 1$ and $\gamma_{ij} \rightarrow 0$ for each edge.  The proportions will determine the sampling probability, and sampling each $\gamma_{ij}$  can be viewed as achieving a \textit{biased} estimator of $-\log p(Y)$.


\vspace*{-0.2cm}
\subsection{From Edge Uncertainty to Graph Attention}
\vspace*{-0.1cm}
Generally speaking, the added flexibility of (\ref{eq:sparse_graph_prior}) comes with a significant cost in terms of algorithmic complexity.  In particular, sampling-based approximations notwithstanding, the integral over each $\gamma_{ij}$ may be intractable, and the resulting MAP problem is generally a nonconvex analogue to (\ref{eq:basic_objective}) with no closed-form solution.  Interestingly though, closer inspection of (\ref{eq:sparse_graph_prior}) allows us to convert the resulting penalty $-\log p(Y)$ into a form directly connected to (\ref{eq:robust_objective}).  More concretely, we can apply the following conversion:

\begin{lemma} \label{lem:gsm_to_convex_bound}
For any $p(Y)$ expressible via (\ref{eq:sparse_graph_prior}), we have
\vspace*{-0.2cm}
\begin{equation} \label{eq:general_penalty}
-\log p(Y) = \pi\left(Y; \rho \right) \triangleq \sum_{\{i,j\} \in \calE} \rho\left( \left\| \by_{i} - \by_{j} \right\|_2^2 \right)
\end{equation}
excluding irrelevant constants, where $\rho : \mathbb{R}^+ \rightarrow \mathbb{R}$ is a concave non-decreasing function that depends on $\mu$.
\end{lemma}
Lemma \ref{lem:gsm_to_convex_bound} can be derived using the framework from \cite{palmer2006variational}; see supplementary for details.  Given that concave, non-decreasing functions dampen the impact of outliers by reducing the penalty applied to increasingly large errors \cite{chen2017strong}, swapping $\pi\left(Y; \rho \right)$ into (\ref{eq:basic_objective}), i.e., as in (\ref{eq:robust_objective}), is a natural candidate for enhancing graph-aware regularization.  However unlike the GNN layers derived previously using (\ref{eq:basic_objective}), it remains unclear how (\ref{eq:robust_objective}), and any subsequent propagation layers built on top of it, relate to existing GNN paradigms or reasonable extensions thereof. 

To address this shortcoming, we introduce a variational decomposition of $\pi\left(Y; \rho \right)$ that allows us to define a family of strict quadratic upper bounds on (\ref{eq:robust_objective}) and ultimately, a useful link to graph attention mechanisms and robust propagation layers.  For this purpose, we first define the approximate penalty function
\vspace*{-0.3cm}
\begin{equation}
\hat{\pi}\left(Y; \widetilde{\rho}, \{\gamma_{ij} \} \right)  \triangleq  \sum_{\{i,j\} \in \calE} \bigg[ \gamma_{ij} \left\| \by_{i} - \by_{j} \right\|_2^2  - \widetilde{\rho}\left(\gamma_{ij} \right) \bigg]
\end{equation}
where $\{\gamma_{ij} \}_{i,j \in \calE}$ denotes a set of variational weights,\footnote{With some abuse of notation, we reuse $\gamma_{ij}$ here as the variational parameters as they are in one-to-one correspondence with edges and function much like a precision variable.} one for each edge, and $\widetilde{\rho}$ represents the concave conjugate of $\rho$ \cite{Rockafellar70}. 
We then form the alternative quadratic energy
\begin{eqnarray} \label{eq:alternative_objective}
&& \hspace*{-0.5cm}  \hat{\ell}_Y(Y;\Gamma, \widetilde{\rho}) ~ \triangleq ~ \left\|Y - f\left(X ; W  \right) \right\|_{\calF}^2 + \lambda \hat{\pi}\left(Y; \widetilde{\rho}, \{\gamma_{ij} \} \right) \nonumber \\
&& \hspace*{-0.2cm} = ~ \left\|Y - f\left(X ; W  \right) \right\|_{\calF}^2 + \lambda \mbox{tr}\left[Y^\top \hat{L} Y  \right] + f\left( \Gamma \right),
\end{eqnarray}
where $\Gamma \in \mathbb{R}^{m \times m}$ is a diagonal matrix with the variational parameter $\gamma_{ij}$ corresponding with each edge arranged along the diagonal,  $f\left( \Gamma \right) \triangleq -\sum_{\{i,j\} \in \calE} \widetilde{\rho}\left(\gamma_{ij} \right)$, and $\hat{L} \triangleq B^\top \Gamma B$.  In this way,  for any fixed set of variational parameters, (\ref{eq:alternative_objective}) is equivalent to (\ref{eq:basic_objective}), excluding constant terms, only now we have the modified Laplacian $\hat{L}$ formed by weighting or attending to the edges $\calE$ of the original graph with $\{\gamma_{ij} \}_{i,j \in \calE}$.

The relevance of this construction then becomes apparent per the following relationships:
\begin{lemma} \label{lem:quadratic_upper_bound}
For all $\{\gamma_{ij} \}_{i,j \in \calE}$, 
\begin{equation} \label{eq:quadratic_upper_bound}
\hat{\ell}_Y(Y;\Gamma, \widetilde{\rho}) ~~ \geq ~~  \ell_Y(Y; \rho),
\end{equation}
with equality\footnote{If $\rho$ is not differentiable, then the equality holds for any $\gamma_{ij}$ which is an element of the subdifferential of $-\rho(z^2)$ evaluated at $z = \left\| \by_{i} - \by_{j} \right\|_2$.} iff
\begin{eqnarray} \label{eq:optimal_solution_gradient}
\gamma_{ij} & = & \arg\min_{\{\gamma_{ij}  > 0 \}} \widetilde{\pi}\left(Y; \widetilde{\rho}, \{\gamma_{ij} \} \right) \\
& = & \left. \frac{\partial \rho\left( z^2 \right) }{\partial z^2}\right|_{z = \left\| \by_{i} - \by_{j} \right\|_2}.  \nonumber
\end{eqnarray}
\end{lemma}
\begin{corollary} \label{cor:optimal_quadratic_bound}
For any $\rho$, there exists a set of attention weights $\Gamma^* \equiv \{\gamma^*_{ij} \}_{i,j \in \calE}$ such that
\begin{equation}
\arg\min_{Y}  \ell_Y(Y; \rho) ~~ = ~~ \arg\min_{Y} \hat{\ell}_Y(Y;\Gamma^*, \widetilde{\rho}).
\end{equation}
\end{corollary}
These results suggest that if we somehow knew the optimal attention weights, we could directly minimize $\ell_Y(Y; \rho)$ using the same efficient propagation rules we derived in Section \ref{eq:basic_unfolding}, only now with $\hat{L} = B^\top \Gamma^* B$ replacing $L$.  But of course $\Gamma^*$ is generally not known in advance, and cannot be computed using (\ref{eq:optimal_solution_gradient}) without known the optimal $Y$.  We describe how to circumvent this issue next.

\subsection{Graph Attention Layers via IRLS}

In prior work involving graph attention, the attention weights are often computed using a (possibly parameterized) module designed to quantify the similarity between nodes sharing an edge, e.g., cosine distance or related \cite{lee2019attention,thekumparampil2018attention}.  While the similarity metric may be computed in different ways, the resulting attention weights themselves are typically formed as $\gamma_{ij} = h(\by_{i},\by_{j})$, where $h$ is a monotonically increasing function of similarity, i.e., more similar nodes receive higher attention weights.  But in general, $h$ is chosen heuristically, as opposed to an emergent functional form linked to an explicit energy.


In contrast, the variational perspective described herein provides us with a natural similarity metric and attention weighting scheme per Lemma \ref{lem:quadratic_upper_bound}.  This ultimately allows us to incorporate attention anchored to an integrated energy function, one that can be directly exploited using attention layers and propagation steps combined together using IRLS.

While IRLS can be derived from many different perspectives, the core formalism is based on a majorization-minimization \cite{hunter2004tutorial} approach to optimization.  In the present context, our goal is to minimize $\ell_Y(Y; \rho)$ from (\ref{eq:robust_objective}).   IRLS operates by alternating between computing an upper bound $\hat{\ell}_Y(Y;\Gamma, \widetilde{\rho})  \geq   \ell_Y(Y; \rho)$ (the \textit{majorization} step), with equality for at least one value of $Y$ as shown in Lemma \ref{lem:quadratic_upper_bound}, and then minimizing this bound over $Y$ (the \textit{minimization} step). 


To begin IRLS, we first initialize $Y^{(0)} = f\left(X ; W  \right)$.  Then until convergence, after the $k$-th iteration we compute:
\begin{enumerate}
    \item Update variational parameters using
    \begin{equation} \label{eq:gamma_update}
    \gamma_{ij}^{(k+1)} ~=~  \left. \frac{\partial \rho\left( z^2 \right) }{\partial z^2}\right|_{z = \left\| \by_{i} - \by_{j} \right\|_2} \equiv ~ h\left(\by_{i}, \by_{j}\right)
    \end{equation}
for all $i,j \in \calE$.  This majorization step updates the bound $\hat{\ell}_Y(Y^{(k)};\Gamma^{(k+1)},\widetilde{\rho})$ and can be viewed as quantifying the similarity between node features $\by_{i}$ and $\by_{j}$ across all edges.  Moreover, because $\rho$ is concave and non-decreasing, the implicit weighting function $h$ so-defined will necessarily be a \textit{decreasing} function of $\left\| \by_{i} - \by_{j} \right\|_2$,  and therefore an \textit{increasing} function of similarity as desired.
    \item Execute one (or possibly multiple) gradient steps on $\hat{\ell}_Y(Y^{(k)};\Gamma^{(k+1)}, \widetilde{\rho})$ via
    \begin{eqnarray} \label{eq:Y_update}
      Y^{(k+1)} &= & Y^{(k)} - \alpha \left. \frac{\partial \hat{\ell}_Y(Y;\Gamma^{(k+1)}, \widetilde{\rho})}{\partial Y} \right|_{\tiny \begin{array}{l} Y = Y^{(k)}, \\ \Gamma = \Gamma^{(k+1)}\end{array}} \nonumber \\
        &=& (1-\alpha) Y^{(k)} + \alpha \left( \tilde{D}^{(k+1)} \right)^{-1} \nonumber  \\
        && \times\left[ \lambda A^{(k+1)} Y^{(k)} + f\left(X ; W  \right) \right], 
    \end{eqnarray}
    where the second line follows from (\ref{eq:basic_grad_step2}), and $A^{(k+1)}$ denotes the adjacency matrix $A$ with edges weighted by $\Gamma^{(k+1)}$; similarly for $\tilde{D}^{(k+1)}$.
And if additional constraints on $Y$ are imposed, then a proximal operator per (\ref{eq:basic_grad_step3}) can also be included here as well.
\end{enumerate}
In aggregate these two steps constitute a single IRLS iteration, with the only caveat being that, while the majorization step (\textit{step 1}) can generally be computed exactly, the minimization step (\textit{step 2}) is only accomplished approximately via movement along the gradient. Even so, we can still guarantee overall cost function descent:
\begin{lemma} 
Provided $\alpha \leq \left\|\lambda B^\top \Gamma^{(k)} B + I\right\|_2^{-1}$, the iterations (\ref{eq:gamma_update}) and (\ref{eq:Y_update}) are such that 
\begin{equation}
\ell_Y(Y^{(k)}; \rho) ~~ \geq ~~ \ell_Y(Y^{(k+1)}; \rho).
\end{equation}
\end{lemma}
This result follows directly from Lemma \ref{lem:quadratic_upper_bound} and properties of IRLS and gradient descent; see supplementary.  

The end result is an iterative algorithm whose individual steps map directly to a rich family of attention-based GNN layers.  To summarize, we need only interleave the propagation rule (\ref{eq:Y_update}) with the attention weight computation from (\ref{eq:gamma_update}).  This ultimately allows us to obtain 
\begin{equation}
    \hat{Y}(W;\rho) ~ \approx ~\arg\min_Y \ell_Y(Y;\rho)
\end{equation}
in such a way that $\partial \hat{Y}(W;\rho)/\partial W$ is actually computable via standard backpropagation.  We may therefore optimize a robust analogue to (\ref{eq:meta_loss}) by executing gradient descent on
\begin{equation} \label{eq:meta_loss2}
\ell_\rho\left(W, \theta  \right) \triangleq \sum_{i=1}^{n'} \calD\bigg( g\left[ \hat{\by}_{i}\left( W; \rho \right); \theta \right], t_i  \bigg)
\end{equation}
to learn optimal parameters $\{W^*,\theta^*\}$.  We refer to this model as TWIRLS, for \textit{Together With IRLS}. The only unresolved design decision then involves the selection of $\rho$.  In this regard, it is helpful to consider a few special cases to elucidate the link with  traditional attention mechanisms.

\subsection{Attention Special Cases}

The function $\rho$ uniquely determines how the attention weights scale with the inverse similarity metric $z_{ij} = \left\| \by_{i} - \by_{j} \right\|_2$.  As a first example, let $\rho\left(z_{ij}^2 \right) = \sqrt{z_{ij}^2} = \left| z_{ij} \right|$, from which it follows using (\ref{eq:optimal_solution_gradient}) that $\gamma_{ij} = \frac{1}{2|z_{ij}|}$ is optimal.  Consequently, based on (\ref{eq:gamma_update}) we have that 
\begin{equation} \label{eq:L1_penalty_weights}
    \gamma_{ij}^{(k+1)} ~=~ \frac{1}{2}\left(\left\| \by^{(k)}_{i} - \by^{(k)}_{j} \right\|_2\right)^{-1}.
\end{equation}
This result implies that if the node embeddings $\by^{(k)}_{i}$ and $\by^{(k)}_{j}$ are very close together, the corresponding value of $\gamma_{ij}^{(k+1)}$ will become large, just as typical attention weights are larger for similar nodes.  Conversely, if $\by^{(k)}_{i}$ and $\by^{(k)}_{j}$ are quite different, then $\gamma_{ij}^{(k+1)}$ will become small, and the resulting quadratic factor in  $\hat{\ell}_Y(Y;\Gamma,\widetilde{\rho})$ will be substantially down-weighted, implying that the corresponding edge could be spurious.

Unfortunately though, (\ref{eq:L1_penalty_weights}) is unbounded from above, and if at any time during training the embeddings along an edge satisfy $\by_{i} \approx \by_{j}$, we have $\gamma_{ij} \rightarrow \infty$.  To avoid this situation, we may instead select $\rho$ from more restricted function classes with bounded gradients.  Table \ref{tab:attention_examples} shows several representative examples along with the corresponding attention weights including upper and lower bounds.

\begin{table}[ht] \caption{Well-behaved robust penalties and their resulting gradients/attention weights. $p \leq 2$, $\epsilon$, and $\tau$ are all non-negative constants. \textit{Range} refers to the allowable attention weights.}
\label{tab:attention_examples}
\begin{center}
\resizebox{\columnwidth}{!}{%
\begin{tabular}{|c|c|c|}
\hline
$\rho\left(z^2\right)$ & $\frac{\partial \rho\left(z^2\right)}{\partial z^2}$ & Range \\ \hline\hline
$\log\left( z^2 + \epsilon \right)$ & $\frac{1}{z^2 + \epsilon}$ & $\left( 0,\tfrac{1}{\epsilon} \right]$ \\ \hline
$ \begin{array}{cc} z^2 &  ~ z < \tau \\ \tau & ~ z \geq \tau\end{array}  $ & $ \begin{array}{cc} 1 &  ~ z < \tau \\ 0 & ~ z \geq \tau \end{array}$ & $\left\{ 0,1 \right\}$ \\ \hline
$ \begin{array}{cc} z^2 &  ~ z < \tau \\ \tau^{(2-p)} z^p & ~ z \geq \tau\end{array}  $ & $ \begin{array}{cc} 1 &  ~ z < \tau \\ \tfrac{p}{2} \tau^{(2-p)}  z^{(p-2)} & ~ z \geq \tau\end{array}$ & $\left( 0,1 \right]$ \\ \hline
\end{tabular} 
}
\end{center}
\end{table}

In general though, there is considerable flexibility in the choice of $\rho$, with different choices leading to different flavors of attention.  And because of the variational link to an explicit energy function associated with different selections, we can directly anticipate how they are likely to behave in various situations of interest.  Note also that in certain cases we may first define a plausible attention update and then work backwards to determine the form of $\rho$ that would lead to such an update.  

\begin{figure}
\centering
    \includegraphics[scale=0.45]{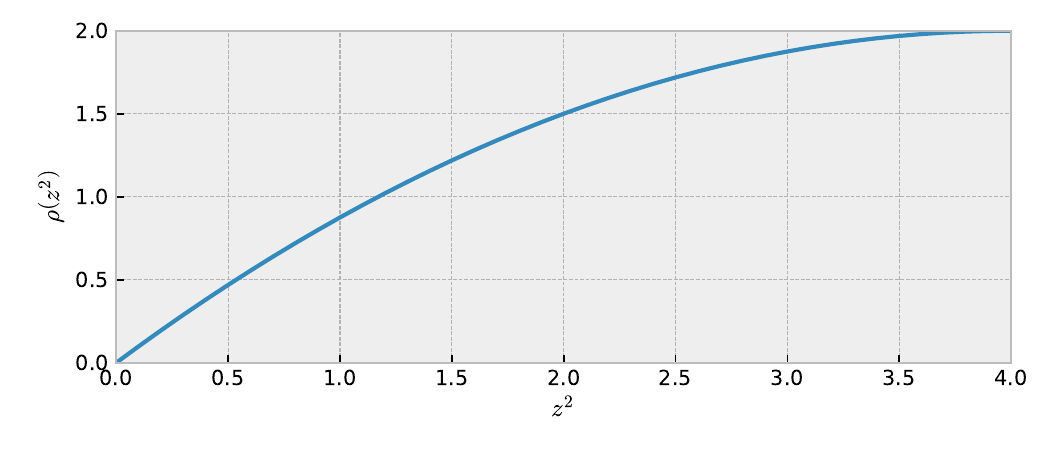}
    \caption{$\rho(z^2)$ vs.~$z^2$ curve that results from using a cosine distance-based attention.}
    \label{fig:cosine-rho}
\end{figure}

For example, the cosine distance is often used as the basis for computing attention weights \cite{lee2019attention,ZhangZ20,thekumparampil2018attention}.  However, if we assume normalized node embeddings $\|\by_{i}\|_2 = 1$ for all $i$, then $\tfrac{1}{2}\left[\cos\angle\left(\by_{i},\by_{j} \right) + 1\right] = 1 - \tfrac{1}{4} \left\|\by_{i} - \by_{j} \right\|_2^2$, which is a non-negative decreasing function of $z_{ij} = \left\|\by_{i} - \by_{j} \right\|_2$, just as the candidate weights produced by TWIRLS will be per Lemmas \ref{lem:gsm_to_convex_bound} and \ref{lem:quadratic_upper_bound}, and as exemplified by the middle column of Table \ref{tab:attention_examples}.  Hence cosine distance could be implemented within TWIRLS, with the requisite unit-norm constraint handled via the appropriate proximal operator.  Additionally, the implicit penalty function $\rho$ that results from using the cosine distance becomes $\rho\left(z^2\right) = z^2 - \frac{1}{8}z^4$, which is illustrated in Figure \ref{fig:cosine-rho}, where we observe the expected concave, non-decreasing design. 

And as a final point of comparison, thresholding attention weights between nodes that are sufficiently dissimilar has been proposed as a simple heuristic to increase robustness \cite{ZhangZ20,Wu0TDLZ19}.  This can also be incorporated with TWIRLS per the second row of Table \ref{tab:attention_examples}, and hence be motivated within our integrated framework.



\section{Related Work} \label{sec:related_work}

Message passing GNNs have been previously designed to handle many of the issues described in Section \ref{sec:intro}, although not generally within a unified framework transparently structured to handle them all.  And in most cases, experiments are limited to the more narrow motivating regime that guided the original design process.  For example, methods developed to reliably extend network depth are typically tested on standard benchmarks with increasing propagation layers \cite{DBLP:conf/icml/ChenWHDL20,li2020deepergcn,DBLP:conf/iclr/RongHXH20,DBLP:conf/icml/XuLTSKJ18}; models that address graph heterophily are tested on datasets with dissimilar labels/features sharing edges \cite{DBLP:conf/nips/ZhuYZHAK20,DBLP:journals/corr/abs-2009-13566,PeiWCLY20}; adversarially-robust designs are deployed against targeted edge or feature purturbations \cite{ZhangZ20,Wu0TDLZ19,zhu2019robust}; and efforts to model long-range dependencies apply graph data where node labels may be sparse \cite{GuC0SG20,DaiKDSS18}. In contrast, we empirically compare TWIRLS across all of these areas in Section \ref{sec:experiments}.  And while some models can be traced back to an underlying energy function \cite{klicpera2019predict,jia2020residual}, or propagation layers related to gradient descent \cite{zhou2004learning}, TWIRLS is the only architecture we are aware of whereby all components, including propagation layers, nonlinear activations, and attention all explicitly follow from iterations that provably descend a principled objective.

Beyond the above, after our original submission of this work we have also recently become aware of four contemporary references \cite{ma2020unified,pan2021a,thatpaper,zhu2021interpreting} whose foundation, like ours, can be traced back to \cite{zhou2004learning} and the corresponding descent of a quadratic energy as in (\ref{eq:basic_objective}).  But although superficially similar in terms of a common starting point, the subsequent analysis in each case is fundamentally different than our derivations and theory from Sections \ref{sec:base_Model} and \ref{sec:IRLS} herein.  In brief, \cite{ma2020unified,pan2021a,thatpaper,zhu2021interpreting} are all focused, at least to some degree, on providing a unified framework for understanding various existing GNN paradigms, whereas we are orthogonally motivated by the robust design of a single new model formed by the merger of proximal gradient descent and IRLS.  

Note also that while the inclusion of attention is loosely discussed in \cite{ma2020unified,pan2021a,zhu2021interpreting}, none of these works actually account for a fully integrated attention mechanism whereby the attention weights themselves are adaptively computed at each layer on the basis of an overarching energy.  Instead, attention weights are either included as a heuristic add-on to a standard quadratic energy \cite{pan2021a}, or else fixed to a function of the input features $X$ with no adaptation across layers \cite{ma2020unified}.   In contrast, the layer-wise adaptation of TWIRLS attention weights from Section \ref{sec:IRLS} explicitly correspond with descent of the global energy from (\ref{eq:robust_objective}).  Similarly, none of these prior works incorporate nonlinear activations as proximal operators as we have done.  Even so, \cite{ma2020unified,pan2021a,thatpaper,zhu2021interpreting} all provide a useful perspective that is complementary to our own, and in future work it could be beneficial to combine the prescriptions from each.  For example, more refined local neighborhood smoothing, as proposed in \cite{pan2021a} based on the similarity of initial input features (as opposed to layer-wise attention), could naturally be incorporated within TWIRLS.

And finally, although the context and motivation is different from our work, GNN architectures have also been previously developed to imitate the individual steps of certain types of graph algorithms \cite{velivckovic2019neural}.  The latter include breadth-first search, Bellman-Ford, and Prim's algorithm. And although not applied to GNN modeling or fully-integrated end-to-end training per se, attention mechanisms have also been derived from an alternative matrix decomposition perspective in  \cite{DBLP:conf/iclr/GengGCLWL21}.



\section{Experiments} \label{sec:experiments}
We test the performance of TWIRLS under different application scenarios, in each case comparing against SOTA methods specifically designed for the particular task at hand.  Across all experiments, we choose either $f\left(X;W \right)$ or $g\left(\by; \theta \right)$ as a shallow MLP (or linear layer), while the other is a simple identity mapping with no parameters.  Hence \textit{all} trainable parameters are restricted to a single module, and TWIRLS operates as an extremely simple architecture. For the attention, we select $\rho$ as a truncated $\ell_p$ quasinorm \cite{wilansky2013modern}, and when no attention is included, we refer to the model as \modelname{base}.  For full details regarding the propagation steps, attention formulation, and other hyperparameters, please see the supplementary; similarly for experimental details, additional ablation studies and empirical results, and a discussion of computational complexity.  All models and experiments were implemented using the Deep Graph Library (DGL) \cite{wang2019dgl}.


\subsection{Base Model Results with No Attention}
We first evaluate \modelname{base} on three commonly used citation datasets, namely Cora, Pubmed and Citeseer \cite{DBLP:journals/aim/SenNBGGE08}, plus ogbn-arxiv \cite{HuFZDRLCL20}, a relatively large dataset. For the citation datasets, we use the semi-supervised setting from \cite{DBLP:conf/icml/YangCS16}, while for ogbn-arxiv, we use the standard split from the Open Graph Benchmark (OGB) leaderboard (see \url{https://ogb.stanford.edu}).  We compare \modelname{base} against the top models from the ogbn-arxiv leaderboard restricted to those with papers and architecture contributions, i.e., we exclude unpublished entries that rely on training heuristics such as adding labels as features, recycling label predictions, non-standard losses/optimization methods, etc.  The latter, while practically useful, can be applied to all models to further boost performance, and are beyond the scope of this work.  Given these criteria, in Table \ref{result-base} we report results for JKNet \cite{DBLP:conf/icml/XuLTSKJ18}, GCNII \cite{DBLP:conf/icml/ChenWHDL20}, and DAGNN \cite{LiuGJ20}. We also include three common baseline models, GCN \cite{kipf2017semi}, SGC \cite{wu2019simplifying} and APPNP \cite{klicpera2019predict}.

\begin{figure}[htbp]
    \centering
    \includegraphics[scale=0.35]{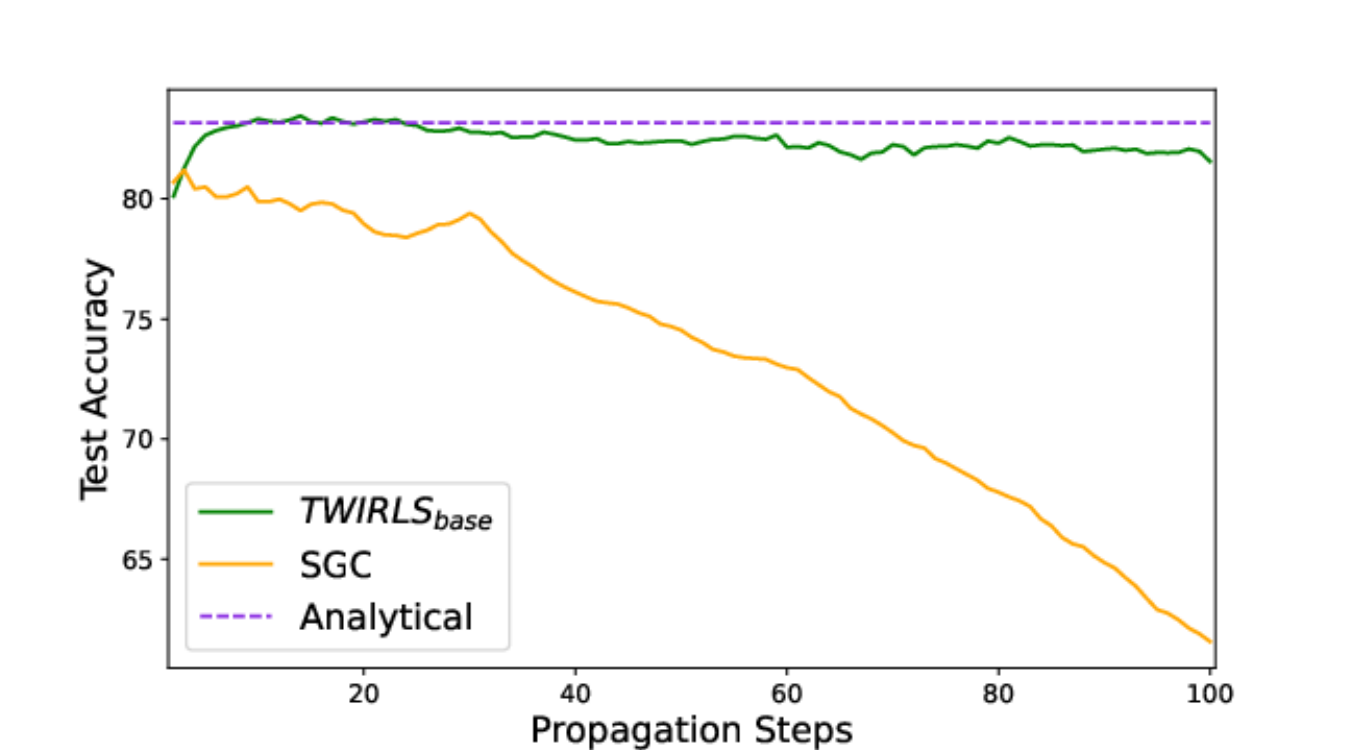}
    \caption{Accuracy versus propagation steps on Cora.}
    \label{vary_prop}
\end{figure}

We also conduct experiments showing the behavior of \modelname{base} as the number of propagation step becomes arbitrarily large. The results on Cora data are plotted in Figure \ref{vary_prop}, where we observe that the performance of \modelname{base} converges to the analytical solution given by (\ref{eq:optimal_L2_solution}) without performance degradation from oversmoothing.  In contrast, the accuracy of SGC drops considerably with propagation.

\vspace{-0.3cm}
\vspace{-0.3cm}

\begin{table}[t] \caption{Baseline results on standard benchmarks.}
\label{result-base}
\begin{center} \begin{scriptsize}\begin{sc}

\begin{tabular}{lcccc}
\toprule
\textbf{Model} & \textbf{Cora} & \textbf{Citeseer} & \textbf{Pubmed} & \textbf{Arxiv}
\\ \midrule
    SGC & 81.7 ± 0.1 & 71.3 ± 0.2 & 78.9 ± 0.1 & 69.79 ± 0.16  \\
    GCN & 81.5 & 71.1 & 79.0 & 71.74 ± 0.29 \\
    APPNP & 83.3 & 71.8 & 80.1 & 71.74 ± 0.29 \\
    JKNet & 81.1 & 69.8 & 78.1 & 72.19 ± 0.21 \\
    GCNII & \textbf{85.5 ± 0.5} & 73.4 ± 0.6 & 80.3 ± 0.4 & 72.74 ± 0.16  \\
    DAGNN & 84.4 ± 0.5 & 73.3 ± 0.6 & 80.5 ± 0.5 & 72.09 ± 0.25  \\ \midrule
    \modelname{base} & 84.1 ± 0.5 & \textbf{74.2 ± 0.45} & \textbf{80.7 ± 0.5} & \textbf{72.93 ± 0.19}  \\
\bottomrule
\end{tabular} 
\end{sc} \end{scriptsize} \end{center} \vskip -0.1in
\end{table}

\begin{table*}[htbp]
\begin{minipage}[ht]{0.65\linewidth}
\begin{center}
\begin{small}
\begin{sc}
\begin{tabular}{lcccc}
\toprule
\textbf{Dataset}         & \textbf{Texas}          & \textbf{Wisconsin}      & \textbf{Actor}          & \textbf{Cornell}        \\
\midrule
Hom.~Ratio ($\calH$)             & 0.11                    & 0.21                    & 0.22                    & 0.3                     \\
\midrule
GCN                    & 59.46$\pm$5.25          & 59.80$\pm$6.99          & 30.26$\pm$0.79          & 57.03$\pm$4.67          \\
GAT                    & 58.38$\pm$4.45          & 55.29$\pm$8.71          & 26.28$\pm$1.73          & 58.92$\pm$3.32          \\
GraphSAGE              & 82.43$\pm$6.14          & 81.18$\pm$5.56          & 34.23$\pm$0.99          & 75.95$\pm$5.01          \\
Geom-GCN               & 67.57                   & 64.12                   & 31.63                   & 60.81                   \\
$\text{H}_2\text{GCN}$ & \textbf{84.86$\pm$6.77} & 86.67$\pm$4.69          & 35.86$\pm$1.03          & 82.16$\pm$4.80          \\
\midrule
MLP                    & 81.89$\pm$4.78          & 85.29$\pm$3.61          & 35.76$\pm$0.98          & 81.08$\pm$6.37          \\
\midrule
\modelname{base}                    & 81.62$\pm$5.51          & 82.75$\pm$7.83          & 37.10$\pm$1.07          & 83.51$\pm$7.30          \\
\modelname{}               & 84.59$\pm$3.83          & \textbf{86.67$\pm$4.19} & \textbf{37.43$\pm$1.50} & \textbf{86.76$\pm$5.05} \\
\bottomrule
\end{tabular}
\end{sc}
\captionof{table}{Result on heterophily graphs.}
\label{table:heterophily-results}
\end{small}
\end{center}
\end{minipage}
\hfill
\begin{minipage}[ht]{0.33\linewidth}
\centering
\includegraphics[width=60mm]{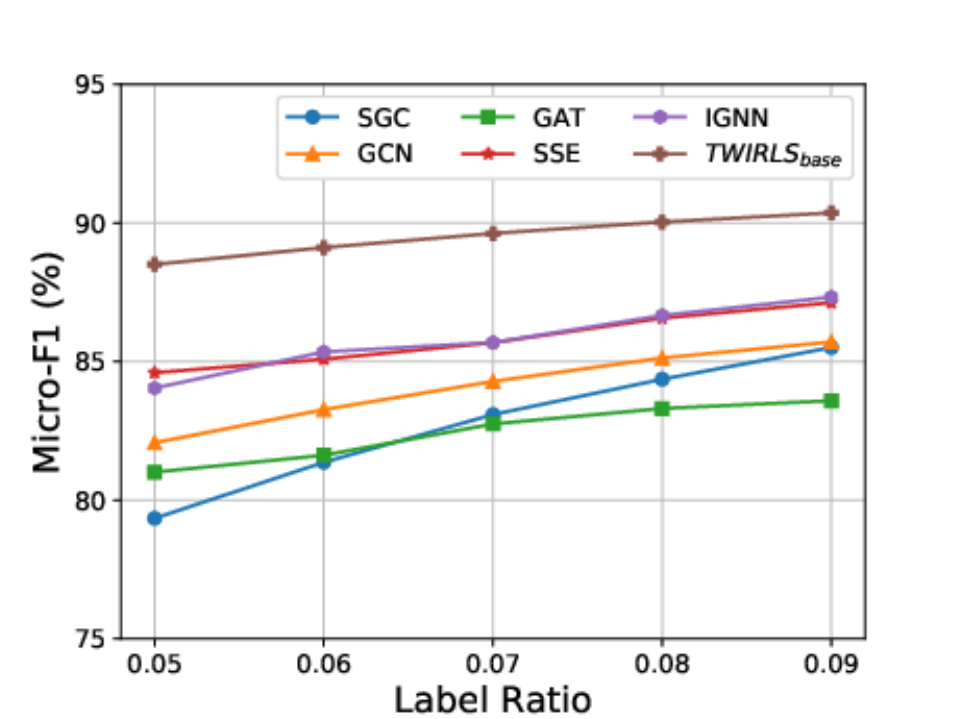}
\captionof{figure}{Amazon co-purchase results.}
\label{fig:amazon-micro}
\end{minipage}
\end{table*}

\subsection{Adversarial Attack Results}
To showcase robustness against edge uncertainty using the proposed IRLS-based attention, we next test the full \modelname{} model using graph data attacked via the the Mettack algorithm  \cite{DBLP:conf/iclr/ZugnerG19}.  Mettack operates by perturbing graph edges with the aim of maximally reducing the node classification accuracy of a surrogate GNN model that is amenable to adversarial attack optimization. And the design is such that this reduction is generally transferable to other GNN models trained with the same perturbed graph.  In terms of experimental design, we follow the exact same non-targeted attack  scenario from \cite{ZhangZ20}, setting the edge perturbation rate to 20\%, adopting the ‘Meta-Self’ training strategy, and a GCN as the surrogate model.  



For baselines, we use three state-of-the-art defense models, namely GNNGuard \cite{ZhangZ20}, GNN-Jaccard \cite{Wu0TDLZ19} and GNN-SVD \cite{EntezariADP20}. Results on Cora and Citeseer data are reported in Table \ref{result-attack}, where the inclusion of attention provides a considerable boost in performance over \modelname{base}.  And somewhat surprisingly, \modelname{} generally performs comparably or better than existing SOTA models that were meticulously designed to defend against adversarial attacks. 

\vspace{-0.2cm}
\begin{table}[ht]
\caption{Adversarial attack comparison with SOTA models.}
\label{result-attack}
 \begin{center} \begin{small} \begin{sc}
\begin{tabular}{lcc}
\toprule
\textbf{Model} & \textbf{ATK-Cora} & \textbf{ATK-Citeseer}
\\ \midrule
    surrogate (GCN) & 57.38 ± 1.42 & 60.42 ± 1.48 \\ \midrule
    GNNGuard & \textbf{70.46 ± 1.03} & 65.20 ± 1.84 \\
    GNN-Jaccard & 64.51 ± 1.35 & 63.38 ± 1.31 \\
    GNN-SVD & 66.45 ± 0.76 & 65.34 ± 1.00 \\ \midrule
    \modelname{base} & 62.89 ± 1.59 & 63.83 ± 1.95 \\
    \modelname{} & 70.23 ± 1.09 &  \textbf{70.63 ± 0.93}\\
\bottomrule
\end{tabular} \end{sc} \end{small} \end{center} \vskip -0.1in
\end{table}

\subsection{Results on Heterophily Graphs}

We also consider non-homophily (or heterophily) graphs to further validate the effectiveness of our IRLS-based attention layers in dealing with inconsistent edges.  In this regard, the homophily level of a graph can be quantified via the homophily ratio $\calH=\frac{|\{(u,v)~:~(u,v)\in\mathcal{E}\land t_u=t_v\}|}{|\mathcal{E}|}$ from \cite{DBLP:conf/nips/ZhuYZHAK20}, where $t_u$ and $t_v$ are the target labels of nodes $u$ and $v$. $\calH$ quantifies the tendency of nodes to be connected with other nodes from the same class. Graphs with an $\calH \approx 1$ exhibit strong homophily, while conversely, those with $\calH \approx 0$ show strong heterophily, indicating that many edges are connecting nodes with different labels.

We select four graph datasets with a low homophily ratio, namely, Actor, Cornell, Texas, and Wisconsin, adopting the data split, processed node features, and labels provided by \cite{PeiWCLY20}. We compare the average node classification accuracy of our model with GCN \cite{DBLP:conf/iclr/KipfW17}, GAT \cite{VelickovicCCRLB18}, GraphSAGE \cite{HamiltonYL17}, SOTA heterophily methods GEOM-GCN \cite{PeiWCLY20} and $\text{H}_2\text{GCN}$ \cite{DBLP:conf/nips/ZhuYZHAK20}, as well as a baseline MLP as reported in \cite{DBLP:conf/nips/ZhuYZHAK20}.

Results are presented in Table \ref{table:heterophily-results}, paired with the homophily ratios of each dataset.  Here we observe that the proposed IRLS-based attention helps \modelname{} generally perform better than \modelname{base} and existing methods on heterophily graphs, achieving SOTA accuracy on three of the four datasets, and nearly so on the forth.

\subsection{Results on Long-Dependency Data}
Finally we test the ability of our model to capture long-range dependencies in graph data by stably introducing an arbitrary number of propagation layers without the risk of oversmoothing.  For this purpose, we apply the Amazon Co-Purchase dataset, a common benchmark used for testing long-range dependencies given the sparse labels relative to graph size \cite{DaiKDSS18,GuC0SG20}.  We adopt the test setup from \cite{DaiKDSS18,GuC0SG20}, and compare performance using different label ratios.  We include five baselines, namely SGC \cite{DBLP:conf/icml/ChenWHDL20}, GCN \cite{kipf2017semi}, GAT \cite{VelickovicCCRLB18}, SSE \cite{DaiKDSS18} and IGNN \cite{GuC0SG20}. Note that SSE and IGNN are explicitly designed for capturing long-range dependencies.  Even so, our model is able to outperform both of them by a significant margin.

\vspace*{-0.1cm}
\section{Conclusion}
\vspace*{-0.1cm}
In this work we have derived \modelname{}, a simple integrated framework for combining propagation and attention layers anchored to the iterative descent of an objective function that is robust against edge uncertainty and oversmoothing. And despite its generic underpinnings, \modelname{} can match or exceed the performance of many existing architectures, including domain-specific SOTA models tailored to handle adversarial attacks, heterophily, or long-range dependencies.

\section*{Acknowledgements}

Zengfeng Huang is supported by National Natural Science Foundation of China No. 61802069, Shanghai Sailing Program No. 18YF1401200, Shanghai Science and Technology Commission No. 17JC1420200. Zhewei Wei is supported by NSFC No. 61972401, No. 61932001, No. 61832017.




\bibliography{references,wipf_refs,wipf_refs_nips2015_vers}
\bibliographystyle{icml2021}
\clearpage

\appendix

\section{Dataset and Experimental Setting Details}

\textbf{Standard Benchmarks}~ In Section 5.1 of the main paper, we used four datasets, namely Cora, Citeseer, Pubmed and ogb-arxiv. These four are all citation datasets, i.e. their nodes represent papers and edges represents citation relationship. The node features of the former three are bag-of-words. Following \cite{DBLP:conf/icml/YangCS16}, we use a fixed spitting for these three datasets in which there are 20 nodes per class for training, 500 nodes for validation and 1000 nodes for testing. For ogbn-arxiv, the features are word2vec vectors. We use the standard leaderboard splitting for ogbn-arxiv, i.e. papers published until 2017 for training, papers published in 2018 for validation, and papers published since 2019 for testing.

\textbf{Adversarial Attack Experiments}~ As mentioned in the main paper, we tested on Cora and Citerseer using Mettack. We use the DeepRobust library \cite{abs-2005-06149} and apply the exact same non-targeted attack setting as in \cite{ZhangZ20}. For all the baseline results in Table 4, we run the implementation in the DeepRobust library or the GNNGuard official code. Note the GCN-Jaccard results differ slightly from those reported in \cite{ZhangZ20}, likely because of updates in the DeepRobust library and the fact that \cite{ZhangZ20} only report results from a single trial (as opposed to averaged results across multiple trails as we report).

\textbf{ Heterophily Experiments}~ In Section 5.3, we use four datasets introduced in \cite{PeiWCLY20}, among which Cornell, Texas, and Wisconsin are web networks datasets, where nodes correspond to web pages and edges correspond to hyperlinks. The node features are the bag-of-words representation of web pages. In contrast, the Actor dataset is induced from a film-director-actor-writer network \cite{tang2009social}, where nodes represent actors and edges denote co-occurrences on the same Wikipedia page. The node features represent some keywords in the Wikipedia pages. We used the data split, processed node features, and labels provided by \cite{PeiWCLY20}, where for the former, the nodes of each class are randomly split into 60\%, 20\%, and 20\% for train, dev and test set respectively.

\textbf{Long-Range Dependency/Sparse Label Tests}~ In Section 5.4, we adopt the Amazon Co-Purchase dataset, which has previously been used in \cite{GuC0SG20} and \cite{DaiKDSS18} for evaluating performance involving long-range dependencies.  We use the dataset provided by the IGNN repo \cite{GuC0SG20}, including the data-processing and evaluation code, in order to obtain a fair comparison. As for splitting, 10\% of nodes are selected as the test set.  And because there is no dev set, we directly report the test result of the last epoch. We also vary the fraction of training nodes from 5\% to 9\%. Additionally, because there are no node features, we learn a 128-dim feature vector for each node. All of these settings from above follow from \cite{GuC0SG20}.

\textbf{Summary Statistics}~ Table \ref{dataset-info} summarizes the attributes of each dataset.

\begin{table}[htbp] \caption{Dataset statistics. The \textit{FEATURES} column describes the dimensionality of node features. Note that the Amazon Co-Purchase dataset has no node features.}
\label{dataset-info}
\begin{center} \begin{scriptsize} \begin{sc}
\begin{tabular}{lcccc}
\toprule
\textbf{Dataset} & \textbf{Nodes} & \textbf{Edges} & \textbf{Features} & \textbf{Classes}
\\ \midrule
    Cora & 2,708 & 5,429 & 1,433 & 7 \\
    Citeseer & 3,327 & 4,732 & 3,703 & 6 \\
    Pubmed & 19,717 & 44,339 & 500 & 3 \\
    Arxiv & 169,343 & 1,166,243 & 128 & 40 \\ \midrule
    Texas & 183 & 309 & 1,703 & 5 \\
    Wisconsin & 251 & 499 & 1,703 & 5 \\
    Actor & 7,600 & 33,544 & 931 & 5 \\
    Cornell & 183 & 295 & 1,703 & 5 \\ \midrule
    Amazon & 334,863 & 2,186,607 & - & 58 \\
\bottomrule
\end{tabular} \end{sc} \end{scriptsize} \end{center}
\end{table}







\section{Model Specifications}

\subsection{Basic Architecture Design}

The \modelname{} architecture is composed of the input module $f\left(X;W\right)$, followed by the unfolded linear propagation layers defined by (21) interleaved with attention given by (20), concluding with $g(\by;\theta)$.  Note that the attention only involves reweighting the edge weights of the graph (i.e., it does not alter the node embeddings at each layer), and when no attention is included we obtain \modelname{base}.  

\begin{figure}[htbp]
\centering
\includegraphics[width=50mm]{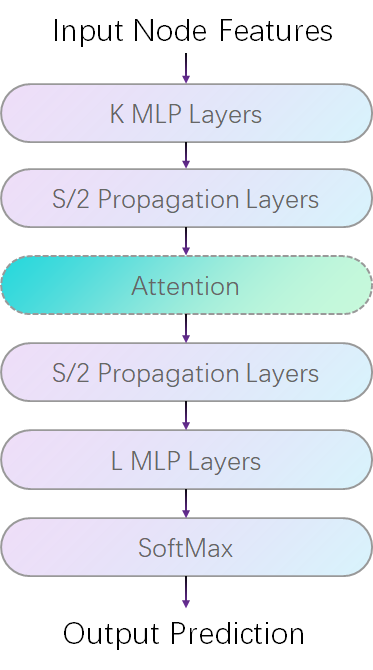}
\captionof{figure}{Model Architecture. S is the total number of propagation steps. K and L are number of MLP layers before and after the propagation respectively. While not a requirement, in all of our experiments, either K or L is set to zero, meaning that the MLP exists on only one side of the propagation layers.}
\label{fig:modelstruct}
\end{figure}

The aggregate design is depicted in in Figure \ref{fig:modelstruct}.  For simplicity, we generally adopt a single attention layer sandwiched between equal numbers of propagation layers; however, for heterophily datasets we apply an extra attention layer before propagation. Additionally, for all experiments except ogbn-arxiv, we set $g(\by;\theta) = \by$ (i.e., an identity mapping).  For ogbn-arxiv, we instead set $f\left(X;W\right) = X$.  Hence for every experiment, \modelname{} restricts all parameters to a single MLP module (or linear layer for some small datasets; see hyperparameter details below).  



\subsection{Specific Attention Formula}
\begin{figure}[htbp]
\begin{center}
\begin{minipage}[ht]{0.65\linewidth}
\begin{center}
\begin{small}
\begin{sc}
\centering
\includegraphics[width=60mm]{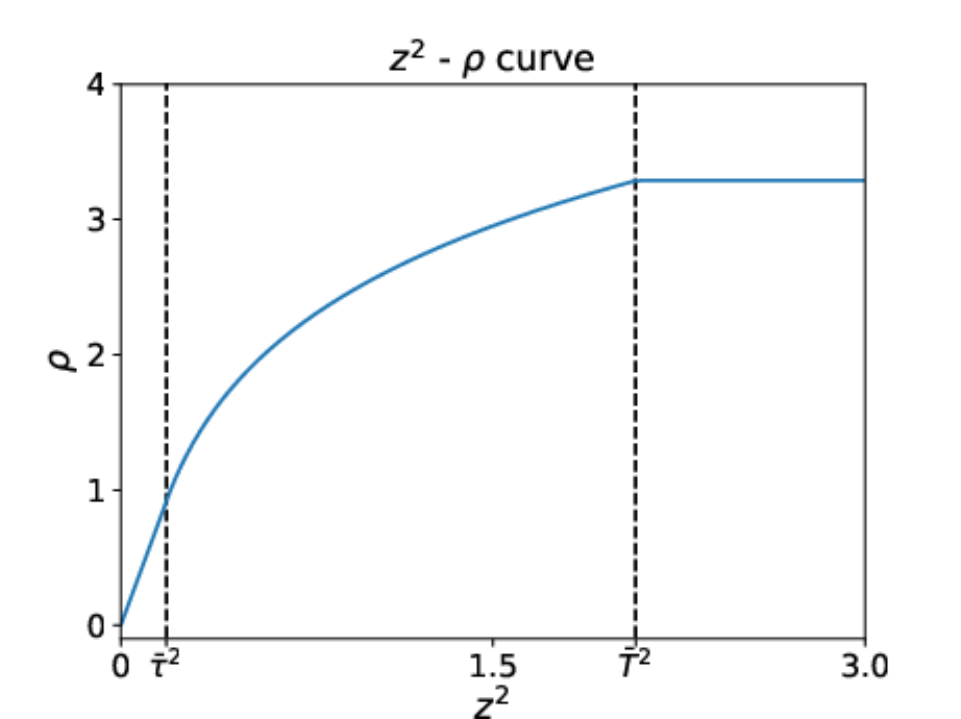}
\end{sc}
\end{small}
\end{center}
\end{minipage}

\begin{minipage}[ht]{0.65\linewidth}
\begin{center}
\begin{small}
\begin{sc}
\centering
\includegraphics[width=60mm]{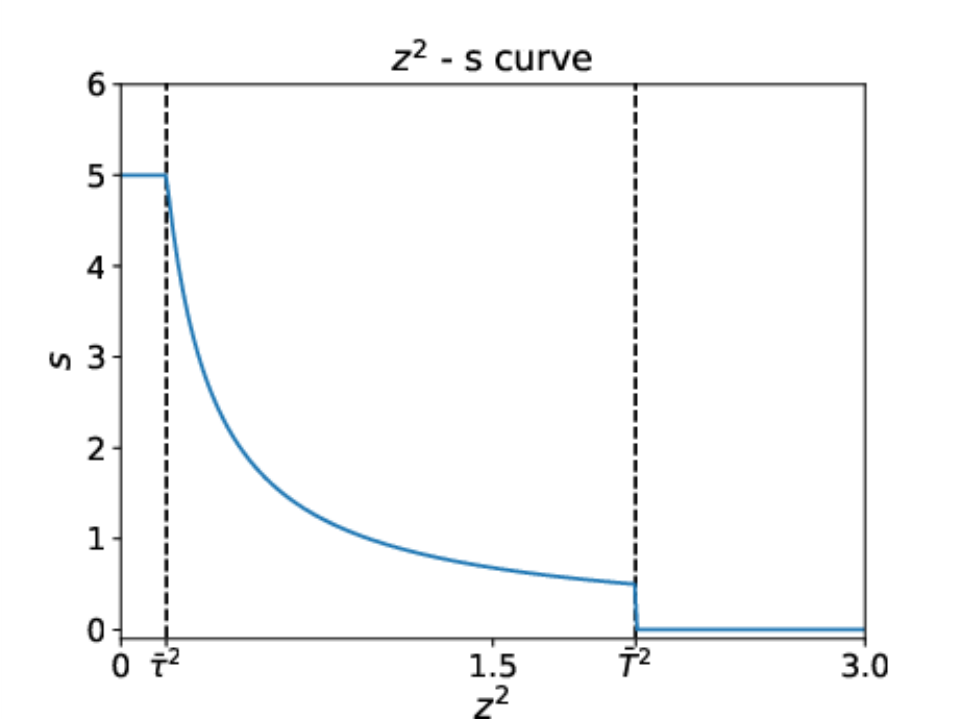}
\end{sc}
\end{small}
\end{center}
\end{minipage}

\captionof{figure}{A visualization of attention functions.}
\label{fig:attn}
\end{center}
\end{figure}

While the proposed attention mechanism can in principle adopt any concave, non-decreasing function $\rho$, in this work we restrict $\rho$ to a single functional form that is sufficiently flexible to effectively accommodate all experimental scenarios.  Specifically, we adopt
\begin{equation}
\rho(z^2) = \begin{cases}\bar \tau^{p-2}z^2 &\quad\text{if }z<\bar \tau \\ \tfrac{2}{p}\bar T^{p} - \rho_0 &\quad \text{if } z>\bar T\\ \tfrac{2}{p} z^{p} - \rho_0 &\quad\text{otherwise},\end{cases}\end{equation} 
where $p$, $\bar T$, and $\bar \tau$ are non-negative hyperparameters and $\rho_0 = \frac{2-p}{p}\bar \tau^p$ is a constant that ensures $\rho$ is continuous. Additionally, the gradient of $\rho$ produces the attention score function (akin to $\gamma$ in the main paper) given by
\begin{equation} s(z^2) \triangleq \frac{\partial p(z^2)}{\partial z^2}  = \begin{cases}\bar \tau^{p-2} &\quad\text{if }z<\bar \tau \\ 0 &\quad \text{if } z>\bar T\\ z^{p-2} &\quad\text{otherwise}.\end{cases}
\end{equation} 
And for convenience and visualization, we also adopt the reparameterizations $\tau = \bar \tau^{\frac{1}{2-p}}$ and $T = \bar T^\frac{1}{2-p}$, and plot $\rho(z^2)$ and $s(z^2)$ in Figure \ref{fig:attn} using $p=0.1$, $\tau=0.2$, $T=2$.                     




Overall, this flexible choice has a natural interpretation in terms of its differing behavior between the intervals $[0,\bar{\tau}]$, $(\bar{\tau},\bar{T})$, and $(\bar{T},\infty)$.  For example, in the $[0,\bar{\tau}]$ interval a quadratic penalty is applied, which leads to constant attention independent of $z$.  This is exactly like \modelname{base}.  In contrast, within the $(\bar{T},\infty)$ interval $\rho$ is constant and the corresponding attention weight is set to zero (truncation), which is tantamount to edge removal.  And finally, the middle interval provides a natural transition between these two extremes, with $\rho$ becoming increasingly flat with larger $z$ values.

Additionally, many familiar special cases emerge for certain parameter selections.  For example $T=\infty$ corresponds with no explicit truncation, while $p=2$ can instantiate no attention.  And $T=\tau$ means we simply truncate those edges with large distance, effectively keeping the remaining edge attention weights at $1$ (note that there is normalization during propagation, so setting edges to a constant is equivalent to setting them to $1$).


\subsection{Hyperparameters}

All model structure-related hyperparameters for \modelname{base} can be found in Table \ref{table:hyper-params}. For experiments with attention, other hyperparameters of the model are the same as the base version, and additional hyper-parameters introduced by attention can be found in Table \ref{table:attn-hyper-params}.  For training, we use the Adam optimizer for all experiments, with learning rate = 0.1 for Cora, Citeseer, attacked Cora, and Texas, and 0.5 for Pubmed and other heterophily datasets. For ogbn-arxiv and Amazon Co-Purchase, we use learning rate = 1e-3 and 1e-2 respectively. 

\begin{table}[htbp]

\begin{center} \begin{scriptsize}\begin{sc}
\begin{tabular}{lccccc}
\toprule
\textbf{\textsc{Dataset}} & $\begin{array}{c} \textbf{\textsc{\# prop}} \\ \textbf{\textsc{layers}} \end{array}$  & \textbf{\textsc{$\lambda$}} & \textbf{\textsc{$\alpha$}} & $\begin{array}{c} \textsc{MLP} \\ \textsc{layers} \end{array}$  & $\begin{array}{c} \textsc{hidden} \\ \textsc{layer} \\ \textsc{size}  \end{array}$  \\
\midrule
\textsc{Cora}         & 16 & 1      & 1    & 2  & -       \\
\textsc{Citeseer}     & 16 & 1      & 1    & 2  & -       \\
\textsc{Pubmed}       & 40 & 1      & 1    & 1  & -       \\
\textsc{Arxiv}        & 7  & 20     & 0.05 & 3  & 512   \\ \midrule

\textsc{ATK-Cora}     & 32 & 1      & 1    & 1  & -       \\
\textsc{ATK-Cite} & 64 & 1      & 1    & 1  & -       \\ \midrule

\textsc{Wisconsin}    & 4  &  0.001 & 1    & 2  & 64     \\
\textsc{Cornell}      & 4  &  0.001 & 1    & 2  & 64     \\
\textsc{Texas}        & 6  &  0.001 & 1    & 2  & 64     \\
\textsc{Actor}        & 6  &  0.001 & 1    & 2  & 64     \\ \midrule

\textsc{Amazon}       & 32  & 10    & 0.1  & 1  & 128   \\

\bottomrule
\end{tabular}
\captionof{table}{Model hyperparameters for \modelname{base}.}
\label{table:hyper-params}
\end{sc} \end{scriptsize} \end{center}  
\end{table}

Consistent with prior work, we apply L2 regularization on model weights, with the corresponding weight decay rate set to 5e-4 for Cora, Pubmed, Wisconsin and Texas, 1e-3 for Citeseer, attacked citeseer, Cornell and Actor, 5e-5 for attacked Cora and 0 for others. Again, as in prior work, we also used dropout as regularization, with dropout rate set to 0.8 for Cora and Pubmed, 0.5 for Citeseer, ogb-arxiv, attcked Cora and attacked Citeseer, and 0 for other datasets.

\begin{table}[htbp]
\begin{center}
\begin{sc}
\begin{tabular}{lccc}
\toprule
\textbf{Dataset} & $p$ & $\tau$ & $T$  \\
\midrule

ATK-Cora     & 0.1 & 0.2 & 2\\
ATK-Citeseer & 0.1 & 0.2 & 2 \\ \midrule

wisconsin    & 1 & 0.1   & $+\infty$ \\
cornell      & 0 & 0.001 & $+\infty$\\
texas        & 0 & 10    & $+\infty$\\
geom-film    & 1 & 0.1   & $+\infty$\\ 

\bottomrule
\end{tabular}
\end{sc}
\captionof{table}{Attention Hyperparameters for \modelname{}}
\label{table:attn-hyper-params}
\end{center}
\end{table}

\section{Model Variations}

\subsection{Alternative GCN-like Reparameterization} \label{sec:gcn_connection_supp}

If we define the reparameterized embeddings $Z = \tilde D^{1/2}Y$ and left multiply (6) by $\tilde D^{1/2}$, we have
{\small\begin{align}Z^{(k+1)} & = (1-\alpha)Z^{(k)} + \alpha\lambda\tilde D^{-1/2}AY^{(k)} + \alpha \tilde D^{-1/2}f(X;W) \nonumber
\\ & = (1-\alpha)Z^{(k)} + \alpha\lambda\tilde D^{-1/2}A\tilde D^{-1/2}Z^{(k)} + \alpha \tilde D^{-1}Z^{(0)}. \label{eq-prop-1}
\end{align}} From here, if we choose $\alpha = \lambda = 1$, for $Z^{(1)}$ we have that 
\begin{eqnarray}
Z^{(1)} & = & \left(\tilde D^{-1/2}A\tilde D^{-1/2} + \tilde D^{-1}\right)Z^{(0)} \nonumber \\
& = & \tilde D^{-1/2}\tilde A\tilde D^{-1/2}Z^{(0)},
\end{eqnarray}
which gives the exact single-layer GCN formulation in $Z$-space with $Z^{(0)} = f(X;W)$.

\subsection{Normalized Laplacian Unfolding}
From another perspective, if we replace $L$ in (1) with a normalized graph Laplacian, and then take gradients steps as before, there is no need to do preconditioning and reparameterizing. For example, following (5) with $L$ changed to the symmetrically-normalized version $\tilde L = I - \tilde D^{-1/2}\tilde A\tilde D^{-1/2}$, we get{\small\begin{equation}Y^{(k+1)} = (1-\alpha-\alpha\lambda)Y^{(k)} + \alpha\lambda \tilde D^{-1/2}\tilde A \tilde D^{-1/2}Y^{(k)} + \alpha Y^{(0)},\label{eq-prop-2}
\end{equation}}
where we set $\tilde D = I+D$. This formula is essentially the same as (\ref{eq-prop-1}). The main difference is that there is no $\tilde D^{-1}$ in front of $X$, which indicates an emphasis on the initial features. We found this version to be helpful on ogbn-arxiv and Amazon Co-Purchase data.  Note however that all of our theoretical support from the main paper applies equally well to this normalized version, just with a redefinition of the gradient steps to include the normalized Laplacian.


\subsection{Layer-Dependent Weights}
It is also possible to seemlessly address the introduction of layer/iteration-dependent weights.  Within the unfolding framework, this can be accomplished by simply changing the specification of the norms used to define $\ell_Y(Y)$. For example, if at each iteration we swap the stated parameter-free Frobenius norms with the reweighted alternative $\| U \|^2_{W_f^s} = \mbox{trace}\left[ U^\top W_f^s U \right]$, where $W_f^s$ is a learnable PSD matrix, then the initial residual term $(1-\alpha)Y^{(k)}$ in (6) will be replaced by $\left(I-\alpha W_f^s\right)Y^{(k)}$. Similarly, if we replace the trace term in (1) by $\mathrm{tr}\left[Y^\top L Y W_p^s\right]$ where $W_p^s$ is a symmetric learnable matrix, then the $AY^{(k)}$ term in (6) would be replaced by $W_p^sAY^{(k)}$, which includes shared learnable parameters.

While this additional flexibility may at times be useful, in the interest of simplicity, for the experiments presented herein we did not include them. More detailed discussion of this general form of energy function and the impact of the symmetry constraint of the parameter matrix is provided in \cite{DBLP:journals/corr/abs-2111-06592}.


\subsection{Sampling and Spectral Sparsification}
The recursive neighborhood expansion across layers poses time and memory challenges for training large and dense graphs.  In this regard, sparsifying the graph in each layer by random sampling is an effective technique, which significantly reduces training time and memory usage but still allows for competitive accuracy \cite{chen2018fastgcn,chen2018stochastic}. Edge sampling has also proved to be effective for relieving over-fitting and over-smoothing in deep GCNs \cite{rong2019dropedge}. 

More specifically, the normalized adjacency matrix $\tilde{A}$ can be sparsified via random sampling.  Let $\tilde{A}'$ denote this sparse version. Then  
\begin{equation}
Z'^{(k+1)} = \mathsf{ReLU}(\tilde{D}^{-1/2}\tilde{A}'\tilde{D}^{-1/2} Z'^{(k)})
\end{equation}
represents the corresponding GCN embedding update.  Note that the sparse random matrix $\tilde{A}'$ used in each layer will generally be different i.i.d.~samples. 
It should also be observed that nonlinear activation functions make the overall function rather complicated, and in particular, it is difficult to get unbiased estimators, i.e., $Z^{(k)} = \E \left[ Z'^{(k)} \right]$ for all $k$. To address this issue, \cite{chen2018fastgcn} assume that there is a (possibly infinite) graph $\calG'$ with the
vertex set $\calV'$ associated with a probability space $(\calV', \calF, \calP)$, such that for the given graph $\calG$, it is an induced subgraph of $\calG'$ and its vertices are i.i.d.~samples of $\calV'$ according to the probability measure $\calP$. But even granted this strong assumption, \cite{chen2018fastgcn} were only be able to show that $Z'$ is a consistent estimator of $Z$, but not necessarily unbiased. Consequently, it can be argued that the theoretical foundation of why random sampling does not significantly impact the accuracy still remains at least partially unclear.

In this context, the perspective on GCNs from Sections 2.2 (main paper) and \ref{sec:gcn_connection_supp} (supplementary) provides a simple alternative explanation. 
Let $L'$ be the Laplacian matrix of the subsampled graph with appropriate scaling such that we have $L=\E[L']$. And let $\ell'_Y(Y) =  \|Y - f(X;W)\|_{\calF}^2 + \lambda \tr \left(Y^TL'Y\right)$. It is then easy to check that $\ell_Y(Y) = \E[\ell'_Y(Y)]$ for all $Y$. And givin the embeddings  
\begin{align}
Z &= \arg\min_{Y} \ell_Y(Y), \textrm{ s.t. } Y\ge 0 \textrm{ and} \nonumber \\
Z' & = \arg\min_{Y} \ell'_Y(Y), \textrm{ s.t. } Y\ge 0,
\end{align}
we observe that, even though $Z'$ is not an unbiased estimator of $Z$, it is nonetheless the optima of an objective function $\ell'_Y(Y)$ that is an unbiased estimator of the corresponding objective for $Z$.

Additionally, per this interpretation, we can also apply spectral sparsification results to get strong theoretical guarantees on graph sparsification for GCNs. For dense graphs with $n$ vertices and $m$ edges, we have that $m=\Omega(n^2)$, which is huge for moderately large graphs. However, it has been proven that there exists a sparse graph $\calG'$, with the same set of vertices and with its edge set a reweighted subset of $\calE$, satisfying 
\begin{equation}
    x^TL'x =(1\pm \varepsilon) x^TLx \textrm{ ~~~~for all $x$},
\end{equation}
where $L'$ is the Laplacian of $\calG'$, and the number of edges in $\calG'$ is $O\left(\frac{n}{\varepsilon^2}\right)$ \cite{batson2012twice}. Moreover, $\calG'$ can be computed in near-linear time \cite{lee2017sdp}. From this result and our interpretation of GCNs, we can always obtain a sparse GCN with constant number of edges per vertex, which approximates the GCN defined by the original graph well. 
This provides nontrivial theoretical guarantees for graph sparsification for graph neural networks.

\section{Ablation Study}

\subsection{Varying $\alpha$ and \# of Propagation Steps}

In the main paper, we interpret $\alpha$ as the gradient step size. From this viewpoint, if the step size becomes smaller, we might naturally expect that more steps are needed for the model to obtain good performance.  To verify this interpretation, we vary $\alpha$ and the number of propagation steps $S$ on Citeseer and observe the performance of \modelname{base}. The results are shown in Table \ref{table:citeseer-grid}.  In general, the best results are arranged on a counter diagonal, which indicates a matching of $\alpha$ and the number of propagation steps gives the best result as expected.

\begin{table}[htbp]
\begin{center}
\begin{sc}
\begin{tabular}{l|cccc}
\toprule
 \diagbox{$S$}{$\alpha$} & 0.1 & 0.25 & 0.5 & 1  \\
\midrule

    8     & 66.00 & 67.25 & 69.51 & 72.35 \\
    16    & 66.80 & 69.23 & 72.56 & \textbf{74.07} \\
    32    & 68.71 & 72.55 & \textbf{74.00} & 73.78 \\
    64    & \textbf{71.66} & \textbf{73.98} & 73.84 & 72.58 \\

\bottomrule
\end{tabular}
\end{sc}
\captionof{table}{\modelname{base} performance on Citeseer as the step size $\alpha$ and the number of propagation steps $S$ are varied. Note that for computational efficiency, the number of repeated experiments here is lower than that of the main paper, so the results are slightly different (we also omit standard deviations for compactness).}
\label{table:citeseer-grid}
\end{center}
\end{table}

\subsection{Varying Truncation Parameter $T$}

We also show how results change when using a different truncating hyperparameter $T$ on attacked Cora and Citeseer.  The results are reported in Table \ref{table:truncating}. Overall, the model performance is relatively stable with respect to $T$, with the best performance occurring with $T=2$.


\begin{table}[htbp]
\begin{center} \begin{threeparttable} 
\begin{sc}
\begin{tabular}{l|cc}
\toprule
 \diagbox{$T$}{data} & ATK-Cora & ATK-Citeseer  \\
\midrule

    $\tau$    & 70.10 ± 1.03  & 69.56 ± 1.32  \\
    2         & \textbf{70.23 ± 1.09}  & \textbf{ 70.63 ± 0.93} \\
    $+\infty$ & 69.77 ± 1.26  & 70.51 ± 1.09   \\

\bottomrule
\end{tabular}
\end{sc}
\captionof{table}{\modelname{} performance under adversarial attacks with different $T$.}
\label{table:truncating} \end{threeparttable}
\end{center}
\end{table}

\subsection{Varying MLP Layers}

In Table \ref{table:multilayer-mlp} we demonstrate that even with only one MLP layer (which is linear), the performance of \modelname{base} is still competitive. 

\begin{table}[htbp]
\begin{center} \begin{threeparttable}
\begin{sc}
\begin{tabular}{l|ccc}
\toprule
 \diagbox{$K$}{data} & Cora & Citeseer  & Pubmed\\
\midrule

    1    & 83.3 ± 0.3          & 74.1 ± 0.5            & \textbf{80.7 ± 0.5} \\
    2    & \textbf{84.1 ± 0.5} & \textbf{74.2 ± 0.45}  & \textbf{80.7 ± 0.4}\\

\bottomrule
\end{tabular}
\end{sc}
\captionof{table}{\modelname{base} performance with different number of MLP layers before propagetion. Here $K$ denotes the number of MLP layers before propagation.}
\label{table:multilayer-mlp} \end{threeparttable}
\end{center}
\end{table}
\section {Additional Empirical Results}

\subsection{Running Time}
The time complexity of our model is $O(mdS+Nd^2)$, where $m$ is the number of edges, $S$ is the number of propagation steps, $N$ and $d$ are the number of MLP layers and hidden size respectively. By contrast, the time complexity of a GCN is $O(mdN+Nd^2)$. Moreover, if the MLP layers are all after propagation (i.e., no parameters before propagation), the time complexity can be reduced to $O(Nd^2)$ by precomputing the propagation (i.e., the same as an MLP). 

To examine empirically, we pick three ogbn-arxiv SOTA models, as well as common baselines GCN, GAT, and MLP (no graph).  We train each for 100 epochs on ogbn-arxiv with a single Tesla T4 and report the average time per epoch in Table \ref{tab:timing} (in seconds).  For consistency, all models have three hidden layers and three propagation layers, and we adjust the hidden size so that the total number of parameters is roughly equivalent for all models. \modelname{}$^*$ denotes that MLP layers are after propagation; otherwise they are before propagation.  Note that for the ogbn-arxiv experiment from Table 2  we use $\text{TWIRLS}^*_\text{base}$, so the computational cost is negligibly different from an MLP, i.e., both are $O(Nd^2)$. 


\begin{table}
\caption{Running time.}
\label{tab:timing}
\begin{center} \begin{sc} 
\begin{tabular}{lcccc}
\toprule
Model & Train  & Test  & \# Parameters \\ \midrule
MLP   & 0.672 & 0.124 & 351,272 \\
\midrule
GCN   & 0.775 & 0.192 & 351,272 \\
GCNII & 0.839 & 0.304 & 317,776 \\
JKNet & 0.998 & 0.285 & 362,920 \\
DAGNN & 0.769 & 0.154 & 351,313 \\
\modelname{base} & 0.746 & 0.169 & 351,272 \\
$\text{TWIRLS}^*_\text{base}$ & 0.679 & 0.124 & 351,272  \\ \midrule
GAT   & 1.486 & 0.266 & 352,336 \\
\modelname{} & 0.814 & 0.189 & 351,272 \\
$\text{TWIRLS}^*$ & 0.775 & 0.223 & 351,272 \\    \bottomrule
  \end{tabular}    \end{sc} \end{center}
\end{table}

\subsection{Amazon Co-Purchase}

\begin{figure}[htbp]
\centering
\includegraphics[width=60mm]{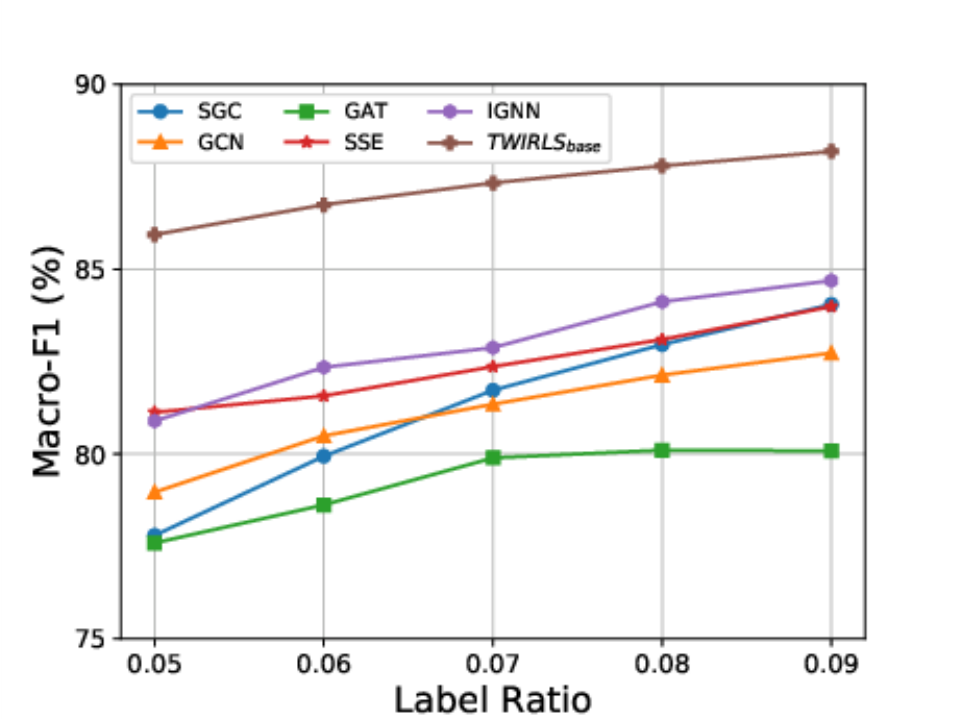}
\captionof{figure}{Amazon Co-Purchase Macro-F1 results.}
\label{fig:amazon-macro}
\end{figure}

In Figure 3 from the main paper, we show the Micro F1 performance of our model on the Amazon Co-Purchase dataset. Here in Figure \ref{fig:amazon-macro} we present the corresponding Macro F1 curve to provide a more detailed picture of our model's ability to capture long-range dependencies.

\subsection{Chains Long-Range Dependency Dataset}

\begin{figure}[htbp]
\centering
\includegraphics[width=60mm]{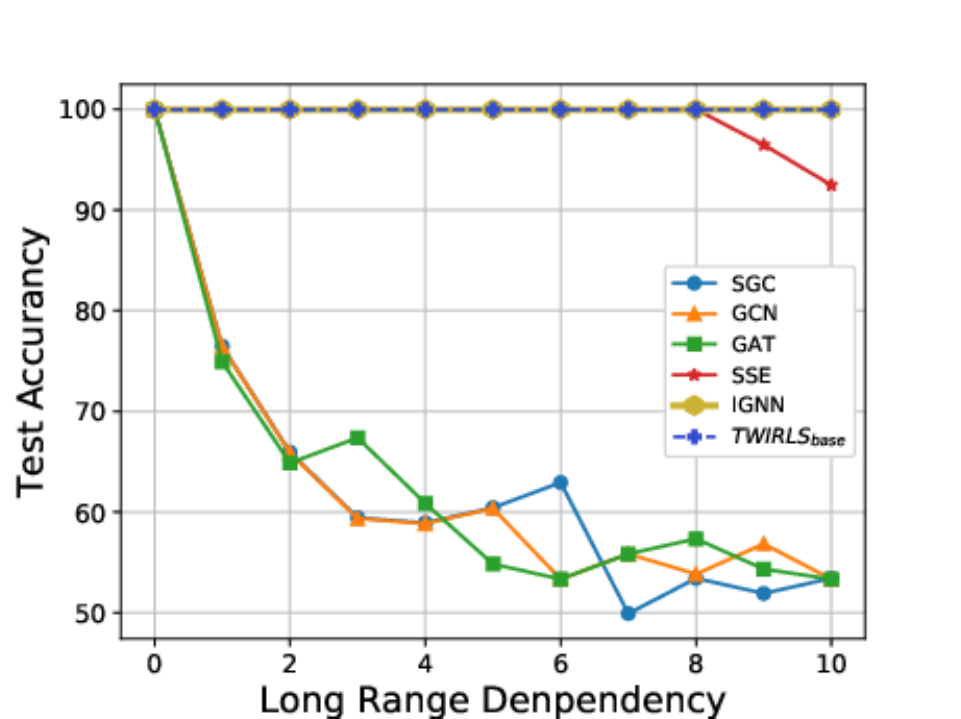}
\captionof{figure}{Results on Chains dataset.  \modelname{base} achieves 100\% accuracy as the long-range dependency is increased by varying chain lengths.}
\label{fig:chains-result}
\end{figure}

To further showcase the ability of our model to capture long-range dependencies, we tested \modelname{} using the Chains dataset introduced by  \cite{GuC0SG20}.  Note that this data has been explicitly synthesized to introduce long-range dependencies of controllable length.  This is accomplished by constructing a graph formed from several uncrossed chains, each randomly labeled 0 or 1.  There is also a 100-dim feature for each node.  And for the node at one end of the chain, the first dimension of its feature vector is the label of this chain; for other nodes the feature vector is a zero vector.  See \cite{GuC0SG20} for further details.  Figure \ref{fig:chains-result} reveals that our model can achieve 100\% accuracy on this data, unlike several of the baselines reported in \cite{GuC0SG20}.  Again, \modelname{} was not designed for this task, but nonetheless performs well.


\subsection{Chinese Word Segmentation} 

Finally, we also observed that \modelname{base} is even able to capture long-range dependencies contained within sentences when we model each sentence as a chain.  To demonstrate this capability, we test our model on a Chinese word segmentation (CWS)  dataset, namely the so-called PKU dataset \cite{DBLP:conf/acl-sighan/Emerson05}. Here each sample is a Chinese sentence with a label for each character, indicating whether this character is the start of a word, middle of a word, end of a word, or itself forms a word. The model then needs to predict the correct labels.

By viewing each Chinese character as a node, and connecting an edge between neighboring characters, we construct a graph from a sentence in a relatively naive way. We use a static character embedding trained by FastNLP\footnote{https://github.com/fastnlp/fastNLP} and run \modelname{base} on this graph. We simply set $\alpha=0.5$, $\lambda = 1$, and apply 8 propagation steps, and use a 2-layer MLP after propagation with a hidden size of 512. As baselines, we train a  GCN and MLP with the same number of layers and hidden size. We also include an 8-layer GCN (denoted by GCN-8), to allow the GCN to have more propagation steps.  And as an additional baseline explicitly designed for modeling dependencies within text sequences, we include a bilateral LSTM that includes one LSTM layer and one linear transform layer.

Table \ref{table:nlp} shows the resulting Macro F1 scores,  which demonstrates the ability of \modelname{base} to handle sentences reasonably well, while the similar performance of GCN and MLP shows that the former does not have similar ability. Surprisingly, \modelname{base} performance is even comparable with the Bi-LSTM sequence model despite not being designed for this task.  And note also, as a quick preliminary test, we did not tune the hyperparameters, nor finely design the model structure and graph construction for this task, so there are space to further boost performance on this type of task. 

\begin{table}[htbp]
\begin{center}
\begin{sc}
\begin{tabular}{lc}
\toprule
 Model & Test Accuracy\\
\midrule
    
    MLP & 56.68 ± 2.81\\
    GCN & 61.95 ± 0.52\\
    GCN-8 & 37.25 ± 0.62 \\
        \modelname{base} & \textbf{84.86 ± 0.39} \\ \midrule
    bi-LSTM & 90.75 ± 0.52\\

\bottomrule
\end{tabular}
\end{sc}
\captionof{table}{Performance of different models on the CWS task.  \modelname{base} outperforms other graph-based models, and is even competitive with a bilateral LSTM that is explicitly designed to handle long-range dependencies within sequences.}
\label{table:nlp}
\end{center}
\end{table}

\section{Proof of Technical Results}

\textbf{Lemma 3.1}~
\textit{For any} $p(Y)$ \textit{expressible via (13), we have}
\begin{equation} 
-\log p(Y) = \pi\left(Y; \rho \right) \triangleq \sum_{\{i,j\} \in \calE} \rho\left( \left\| \by_{i} - \by_{j} \right\|_2^2 \right) \nonumber
\end{equation}
\textit{excluding irrelevant constants, where} $\rho : \mathbb{R}^+ \rightarrow \mathbb{R}$ \textit{is a concave non-decreasing function that depends on} $\mu$.
\vspace*{0.3cm}

\textbf{\textit{Proof}:}~  A function $f : \mathbb{R}_+ \rightarrow \mathbb{R}_+$ is said to be \textit{totally monotone} \cite{widder2015laplace} if it is continuous on $[0,\infty)$ and infinitely differentiable on $(0,\infty)$, while also satisfying
\begin{equation}
    (-1)^n \frac{\partial^n }{\partial u^n}f(z) ~\geq~ 0, ~~\forall~n = 1,2,\ldots.
\end{equation}
for all $z>0$. Furthermore, a non-negative symmetric function $p_z(z)$ can be expressed as a Gaussian scale mixture, i.e.,
\begin{equation} \label{eq:gsm_supp}
p_z\left(z \right) = \int  \calN\left(z |  0,\gamma^{-1} I \right) d \mu\left(\gamma \right),
\end{equation}
for some positive measure $\mu$, iff $p_z(\sqrt{z})$ is a totally monotone function on $[0,\infty)$ \cite{andrews1974scale}.  However, as shown in \cite{palmer2006variational}, any such totally monotone function can be expressed as $p_z(\sqrt{z}) = \exp\left[-\rho(z) \right]$, where $\rho$ is a concave, non-decreasing function.  From these results, and the assignment $z_{ij} = \sqrt{\bu_{ij}^\top \bu_{ij}} = \|\by_i - \by_j \|_2$, we can then infer that
\begin{eqnarray}
-\log p(Y)  && \nonumber \\
 && \hspace*{-2cm} \equiv -\sum_{\{i,j\} \in \calE} \log \int  \calN\left(\bu_{ij} |  0,\gamma^{-1}_{ij} I \right) d \mu\left(\gamma_{ij}\right) \nonumber \\
&& \hspace*{-2cm}  =   -\sum_{\{i,j\} \in \calE} \log \int \left( \tfrac{\gamma_{ij}}{2\pi} \right)^{d/2} \exp\left[-\tfrac{\gamma_{ij}}{2} \bu^\top_{ij} \bu_{ij}  \right] d \mu\left(\gamma_{ij}\right) \nonumber \\
&& \hspace*{-2cm}  =   -\sum_{\{i,j\} \in \calE} \log \int \left( \tfrac{\gamma_{ij}}{2\pi} \right)^{1/2} \exp\left[-\tfrac{\gamma_{ij}}{2} z_{ij}^2  \right] d \mu'\left(\gamma_{ij}\right) \nonumber \\
&& \hspace*{-2cm} = -\sum_{\{i,j\} \in \calE} \log \int  \calN\left(z_{ij} |  0,\gamma^{-1}_{ij} I \right) d \mu'\left(\gamma_{ij}\right) \nonumber \\
    && \hspace*{-2cm}  =  - \sum_{\{i,j\} \in \calE} \log p_z\left( z_{ij} \right) \nonumber \\
    && \hspace*{-2cm}  =  - \sum_{\{i,j\} \in \calE} \log p_z\left(\sqrt{\|\by_i - \by_j \|^2_2} \right) \nonumber \\
    & & \hspace*{-2cm} = \sum_{\{i,j\} \in \calE} \rho\left( \|\by_i - \by_j \|^2_2 \right),
\end{eqnarray}
where the positive measure $\mu'$ is defined such that $d \mu'(\gamma_{ij}) = \left( \tfrac{\gamma_{ij}}{2\pi} \right)^{(d-1)/2} d \mu(\gamma_{ij})$, noting that prior results apply equally well to this updated version for some concave non-decreasing $\rho$.  Hence Lemma 3.1 directly follows.  \myendofproof

\textbf{Lemma 3.2}~ \textit{For all} $\{\gamma_{ij} \}_{i,j \in \calE}$, 
\begin{equation}
\hat{\ell}_Y(Y;\Gamma, \widetilde{\rho}) ~~ \geq ~~  \ell_Y(Y; \rho), \nonumber
\end{equation}
\textit{with equality}\footnote{If $\rho$ is not differentiable, then the equality holds for any $\gamma_{ij}$ which is an element of the subdifferential of $-\rho(z^2)$ evaluated at $z = \left\| \by_{i} - \by_{j} \right\|_2$.} \textit{iff}
\begin{eqnarray} 
\gamma_{ij} & = & \arg\min_{\{\gamma_{ij}  > 0 \}} \widetilde{\pi}\left(Y; \widetilde{\rho}, \{\gamma_{ij} \} \right) \nonumber \\
& = & \left. \frac{\partial \rho\left( z^2 \right) }{\partial z^2}\right|_{z = \left\| \by_{i} - \by_{j} \right\|_2}.  \nonumber
\end{eqnarray}
\vspace*{0.3cm}

\textbf{Corollary 3.2.1}~ \textit{For any} $\rho$, \textit{there exists a set of attention weights} $\Gamma^* \equiv \{\gamma^*_{ij} \}_{i,j \in \calE}$ \textit{such that}
\begin{equation}
\arg\min_{Y}  \ell_Y(Y; \rho) ~~ = ~~ \arg\min_{Y} \hat{\ell}_Y(Y;\Gamma^*, \widetilde{\rho}). \nonumber 
\end{equation}
\vspace*{0.3cm}

\textbf{\textit{Proof}:}~   Both Lemma 3.2 and Corollary 3.2.1 follow directly from principles of convex analysis and Fenchel duality \cite{Rockafellar70}.  In particular, any concave, non-decreasing function $\rho : \mathbb{R}_+ \rightarrow \mathbb{R}$ can be expressed via the variational decomposition
\begin{eqnarray} \label{eq:basic_variational_form_supp}
    \rho\left(z^2\right) & = & \min_{\gamma > 0} \left[ \gamma z^2 - \rho^*\left( \gamma \right) \right] \nonumber \\
    & \geq & \gamma z^2 - \widetilde{\rho} \left( \gamma \right),
\end{eqnarray}
where $\gamma$ is a variational parameter whose optimization defines the decomposition, and $\widetilde{\rho}$ is the concave conjugate of $\rho$. From a visual perspective, (\ref{eq:basic_variational_form_supp}) can be viewed as constructing $\rho\left(z^2\right)$ as the minimal envelope of a series of quadratic upper bounds, each defined by a different value of $\gamma$.  And for any fixed $\gamma$, we obtain a fixed upper bound once we remove the minimization operator.  By adopting $z = \|\by_i - \by_j\|_2$ for all $i,j \in \calE$ we obtain (16), which by construction satisfies (17).  And (18) follows by noting that at any optimal $\gamma^*$, the upper bound satisfies
\begin{equation}
\gamma^* z^2 - \widetilde{\rho}\left( \gamma^* \right) = \rho\left(z^2\right),
\end{equation}
i.e., it is tangent to $\rho$ at $z^2$, in which case $\gamma^*$ must be equal to the stated gradient (or subgradient).  

And finally, in terms of Corollary 3.2.1, let $Y^* = \arg\min_Y \ell_Y(Y;\rho)$.  We may then simply apply Lemma 3.2 to form the bound
\begin{equation}
\hat{\ell}_Y(Y^*;\Gamma, \widetilde{\rho}) ~~ \geq ~~ \hat{\ell}_Y(Y^*;\Gamma^*, \widetilde{\rho}) ~~ = ~~ \ell_y(Y^*;\rho),
\end{equation}
where $\Gamma^*$ denotes a diagonal matrix with optimized $\gamma_{ij}^*$ values along the diagonal.  Therefore $\hat{\ell}_Y(Y;\Gamma^*, \widetilde{\rho})$ so-defined achieves the stated result.  \myendofproof

\textbf{Lemma 3.3}~ \textit{Provided that} $\alpha \leq \left\|\lambda B^\top \Gamma^{(k)} B +  I\right\|_2^{-1}$, \textit{the iterations (20) and (21) are such that} 
\begin{equation}
\ell_Y(Y^{(k)}; \rho) ~~ \geq ~~ \ell_Y(Y^{(k+1)}; \rho). \nonumber
\end{equation}
\vspace*{0.0cm}

\textbf{\textit{Proof}:}~   We will first assume that no Jacobi preconditioning is used (i.e., $\tilde{D} \equiv I$); later we will address the general case.  Based on Lemma 3.2 and (18), as well as the analogous update rule for $\Gamma^{(k+1)}$ from (20), it follows that
\begin{equation}
\hat{\ell}_Y\left(Y^{(k)};\Gamma^{(k+1)}, \widetilde{\rho}\right) ~~ = ~~  \ell_Y(Y^{(k)}; \rho).
\end{equation}
Now define $\Psi(Y) \triangleq \hat{\ell}_Y\left(Y;\Gamma^{(k+1)}, \widetilde{\rho}\right)$ and
\begin{eqnarray} \label{eq:convex_bound1_supp}
\hat{\Psi}(Y) & \triangleq & \Psi\left(Y^{(k)}\right) +  \\
&& \hspace*{-0.5cm} \nabla \Psi\left(Y^{(k)} \right)^\top\left( Y - Y^{(k)} \right) + \tfrac{\calL}{2}\left\|Y - Y^{(k)} \right\|_{\calF}^2, \nonumber
\end{eqnarray}
where $\Psi$ has Lipschitz continuous gradients with Lipschitz constant $\calL$ satisfying 
\begin{equation} \label{eq:Lipschitz_inequality}
\left\| \nabla \Psi(Y_1) - \nabla \Psi(Y_2) \right\|_{\calF} \leq \calL  \left\|  \Psi(Y_1) - \Psi(Y_2) \right\|_{\calF}
\end{equation}
for all $Y_1$ and $Y_2$.  We may then conclude that
\begin{equation} \label{eq:misc_inequality_supp}
  \hat{\Psi}(Y) ~~ \geq ~~ \Psi(Y) ~~ \geq ~~ \ell_Y (Y;\rho ),
\end{equation}
with equality at the point $Y= Y^{(k)}$.  Note that the first inequality in the above expression follows from basic results in convex analysis (e.g., see \cite{bubeck2014convex}[Lemma 3.4]), while the second comes from (17).  Consequently, by adopting $Y^{(k+1)} = Y^{(k)}-\alpha \nabla\Psi Y^{(k)}$, which is equivalent to (21) excluding the Jacobi preconditioner, we have that
\begin{align}\hat \Psi \left(Y^{(k+1)}\right) & = \hat \Psi\left(Y^{(k)} - \alpha \nabla\Psi Y^{(k)}\right)  \\ & =  \Psi\left(Y^{(k)}\right) +  \left(\frac{\mathcal L \alpha^2}{2}-\alpha\right)\left\|\nabla \Psi\left(Y^{(k)}\right)\right\| \label{eq:descent-iteration} \\ & \leq \Psi\left(Y^{(k)}\right) \label{eq:descent-iteration-2}\end{align}
Note the inequality from (\ref{eq:descent-iteration}) to (\ref{eq:descent-iteration-2}) holds as long as we adopt a step-size $\alpha \leq \tfrac{2}{\calL}$. Hence we must only enforce the Lipschitz constraint (\ref{eq:Lipschitz_inequality}) to guarantee monotonicity.  This is equivalent to the requirement that $\calL I - \nabla^2 \Psi\left(Y^{(k)} \right) \succeq  0$ for all $k$, which computes to $\calL I \succeq 2\left( I + \lambda B^\top \Gamma^{(k)} B\right)$.  Setting $\calL$ to be greater than or equal to the maximum singular value of $2\left(  I + \lambda B^\top \Gamma^{(k)} B\right)$ satisfies this objective, which then leads to the step-size bound $\alpha \leq \left\|\lambda B^\top \Gamma^{(k)} B +  I\right\|_2^{-1}$.

And finally, if we reintroduce the non-trivial Jacobi preconditioner $\tilde{D}^{-1} = \left( \lambda D + I \right)^{-1} \neq I$, we need only redefine the bound from (\ref{eq:convex_bound1_supp}) as
\begin{eqnarray} \label{eq:convex_bound2_supp}
\hat{\Psi}(Y) & \triangleq & \Psi\left(Y^{(k)}\right) +  \nabla \Psi\left(Y^{(k)} \right)^\top\left( Y - Y^{(k)} \right)  \\
&& \hspace*{-0.0cm}  + ~ \tfrac{\calL}{2}\left(Y - Y^{(k)} \right)^\top \left( \tilde{D}^{(k+1)}\right)^2 \left(Y - Y^{(k)} \right).  \nonumber
\end{eqnarray}
And because $\left( \tilde{D}^{(k+1)}\right)^2 \succeq I$, (\ref{eq:misc_inequality_supp}) still holds and the same conclusions follow through as before.  The only major difference is that now
\begin{equation}
    Y^{(k+1)} = Y^{(k)} - \alpha \left(\tilde{D}^{(k+1)}\right)^{-1} \nabla \Psi\left(Y^{(k)} \right),
\end{equation}
which arises from the preconditioned update rule from (21). Then by applying the exact same process as above, we arrive at the conclusion $\Psi\left(Y^{(k+1)}\right) = \hat \Psi\left(Y^{(k+1)}\right) \leq \Psi\left(Y^{(k)}\right)$ as long as $\alpha$ satisfies the stated condition. And in fact, a larger step-size range is actually possible in this situation since the upper bound from (\ref{eq:convex_bound2_supp}) holds for smaller values of $\calL$ (note that all diagonal values of $\tilde{D}^{(k+1)}$ are greater than one).   \myendofproof

    
    
    
    
    


\end{document}